\documentclass[10pt,journal]{IEEEtran}

\usepackage{cite}
\usepackage{amsmath,amssymb,amsfonts}
\usepackage{graphicx}
\usepackage{textcomp}
\usepackage{array}
\usepackage{xcolor}
\usepackage{enumerate}
\usepackage{bm}
\usepackage{algorithm}
\usepackage{booktabs}

\usepackage{multirow}
\usepackage{epstopdf}
\usepackage{amsthm}

\usepackage{graphicx}
\usepackage[subfigure]{graphfig}
\def\BibTeX{{\rm B\kern-.05em{\sc i\kern-.025em b}\kern-.08em
    T\kern-.1667em\lower.7ex\hbox{E}\kern-.125emX}}

\usepackage{enumerate}

\usepackage{algorithm}
\usepackage{algpseudocode}
\usepackage{amsmath}
\begin{document}

\title{TrafficGPT: Towards Multi-Scale Traffic Analysis and Generation with Spatial-Temporal Agent Framework}
\author{
Jinhui~Ouyang,~Yijie~Zhu,~Xiang~Yuan,~Di~Wu,~\IEEEmembership{Member,~IEEE}
\thanks{J. Ouyang, Yi. Zhu, Xiang Yuan, and D. Wu are with the Key Laboratory for Embedded and Network Computing of Hunan Province, Hunan University, Changsha, Hunan 410082, China (e-mail: \{oldyoung, zyj1936013472, yuanxiang, dwu\}@hnu.edu.cn).} 
}
\maketitle

\begin{abstract}
The precise prediction of multi-scale traffic is a ubiquitous challenge in the urbanization process for car owners, road administrators, and governments. In the case of complex road networks, current and past traffic information from both upstream and downstream roads are crucial since various road networks have different semantic information about traffic. Rationalizing the utilization of semantic information can realize short-term, long-term, and unseen road traffic prediction. As the demands of multi-scale traffic analysis increase, on-demand interactions and visualizations are expected to be available for transportation participants. We have designed a multi-scale traffic generation system, namely \textsf{TrafficGPT}, using three AI agents to process multi-scale traffic data, conduct multi-scale traffic analysis, and present multi-scale visualization results. \textsf{TrafficGPT} consists of three essential AI agents: 1) a text-to-demand agent that is employed with Question \& Answer AI to interact with users and extract prediction tasks through texts; 2) a traffic prediction agent that leverages multi-scale traffic data to generate temporal features and similarity, and fuse them with limited spatial features and similarity, to achieve accurate prediction of three tasks; and 3) a suggestion and visualization agent that uses the prediction results to generate suggestions and visualizations, providing users with a comprehensive understanding of traffic conditions. Our \textsf{TrafficGPT} system focuses on addressing concerns about traffic prediction from transportation participants, and conducted extensive experiments on five real-world road datasets to demonstrate its superior predictive and interactive performance. 

\end{abstract}


       




\begin{figure*}[ht]
    \centering
    \includegraphics[width=1\textwidth]{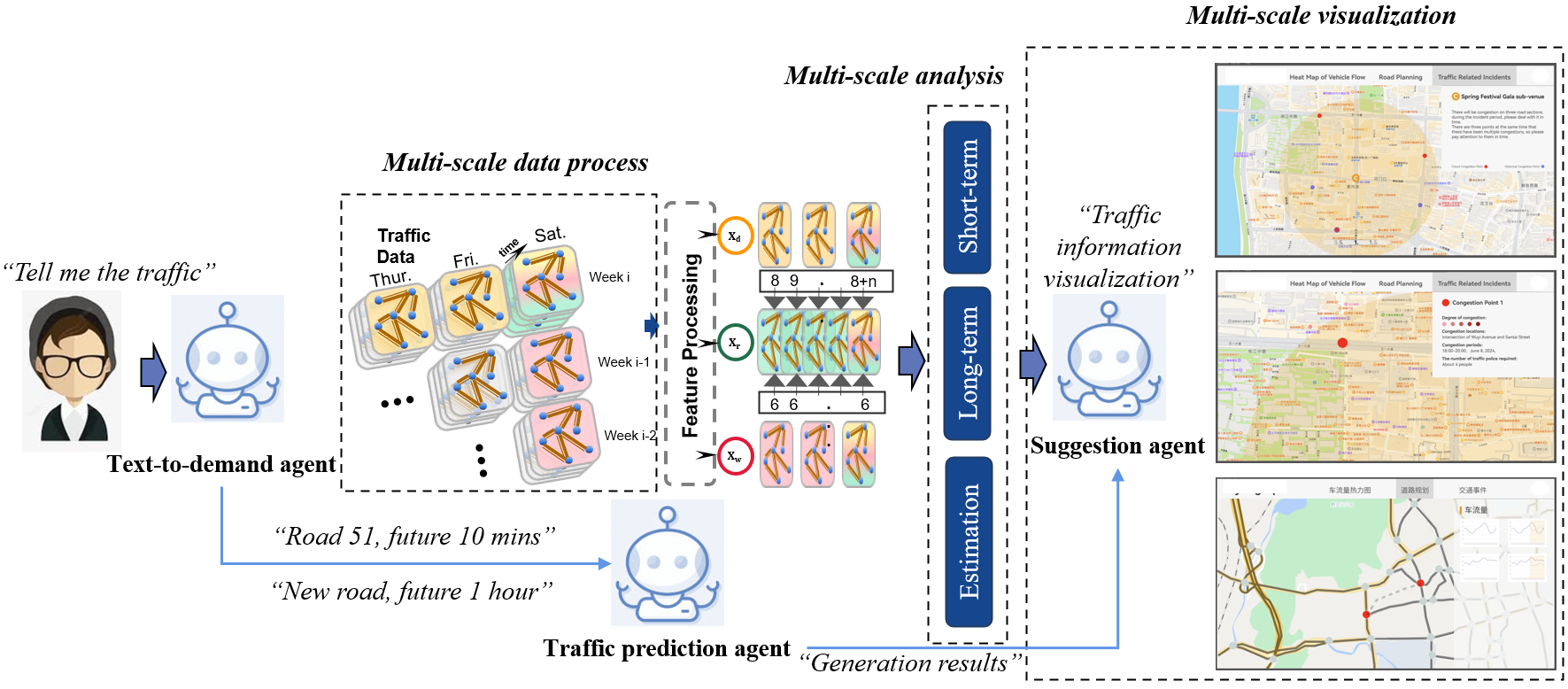}
    \scriptsize
    \caption{TrafficGPT system for multi-scale traffic generation. The system, employed with three agents, uses multi-scale spatial-temporal data to conduct multi-scale traffic analysis and present multi-scale visualization.}
    \label{Abstract}
\end{figure*}
\section{Introduction}
With the urbanization process, road transportation has become more important in cities and has induced an increasing number of vehicles involved in traffic. The increasing number of vehicles has caused road congestion, which is now a significant impediment to the efficiency of modern society~\cite{Jin2020-TITS,Deep2020}. This congestion results in not only higher greenhouse gas emissions but also increased traffic harm. To alleviate traffic congestion, many cities have reconstructed complex road networks. However, construction time and construction distance have a huge impact on urban road congestion~\cite{Wang2016msmi}. Meanwhile, the unpredictable traffic on newly added roads raised concern for car owners~\cite{Kukkapalli2018}. Therefore, rapidly analyzing the new traffic conditions for accurate prediction becomes imperative for transportation participants. At the same time, transportation participants raise increasing demands on innovative and intelligent interaction method for ubiquitous access and multi-scale understanding of traffic data.

Accurate short-term (within 1-hour) prediction of traffic is one of the essential ways to alleviate traffic congestion on urban roads, which have high capacities and large traffic volumes that connect geographically important areas~\cite{2012The,Jin2021}. Nowadays, researchers endeavor to use deep learning to solve this problem. However, existing studies for traffic prediction often rely heavily on spatial feature extraction, which makes it inconvenient to train a deep-learning model for new road networks. Many schemes are proposed to address this issue, for instance, ~\cite{2018CNN,2020CNN_LSTM,2020ConvLSTM,kim2022icde} utilize Convolution Neural Networks (CNNs) to learn temporal features. CNN can also be used for spatial data measured by Euclidean distance\cite{2020TGCN}. Nevertheless, urban road networks are graph-structured data with many spatial features that can be utilized. Recently, Graph Convolutional Networks (GCNs) \cite{2020GCN,YuIJCAI2018} have been proposed to mine the structural features of networks, providing a good solution to the above problems~\cite{RossiSIGMOD2022}. GCN can well capture the static spatial topological relationships of traffic data that have spatial proximity, regional similarity, and spatial heterogeneity. However, GCN struggles to extract the dynamic spatial changes of urban roads, which requires additional research to better capture the dynamical spatial feature. Nowadays, the extraction and use of temporal and spatial features in traffic data are studied for traffic prediction ~\cite{HanKDD2021,Jinhuitits2024}.

Reaching a consensus, traffic prediction is an extensive spatial-temporal study using different traffic data, traffic road scenarios, and prediction durations. Spatial-temporal traffic prediction has been studied in recent years. For example,~\cite{ji2023aaai} proposes a self-learning method to enhance the model, and~\cite{jiang2023aaai, jin2023aaai} propose a new model structure to effectively extract features. However, these existing methods either focus only on the temporal dimension of the data or overamplify the importance of spatial features; the former limits the improvement of prediction accuracy, and the latter makes it inconvenient to train a deep network model for a new road network. Long-term and unseen road traffic predictions with stable accuracy can further alleviate traffic congestion and gradually become a concern. The key challenges to traffic prediction are four-fold:
\begin{enumerate}
\item Most schemes~\cite{zhengaaai2020, YuIJCAI2018,wuijcai2019, renhe2023aaai, guo2021aaai} fail to effectively mine the spatial-temporal dependencies of traffic data, and often prioritize the exploration of intricate spatial features, overlooking the potential of the combination of basic spatial features and temporal features. Under the premise of ensuring accuracy, as few as possible of the spatial features enable the model to be fast-trained and adapt to new road networks.

\item Road networks have rich traffic-semantic information shared by roads, such as the spatial structure that impacts traffic patterns between roads~\cite{Liangtkdd2023}; however, few studies rationalize the use of this information. Effective utilization of this information may contribute to long-term and unseen road prediction tasks. 

\item Traffic data has different scales in time (daily, weekly, monthly, etc.), and it becomes a great challenge to utilize this multi-scale information for multiple prediction tasks. Exploiting spatial-temporal similarity that can be used in multi-scale data processing becomes important.

\item Predictable task growth deteriorates the comprehensibility of the straightforward, monotonous presentation of data. It is important to intelligently capture the predictive tasks corresponding to user needs and improve the comprehensibility of the data.
\end{enumerate}

Noteworthly, with the rapid development of generative Artificial intelligence (AI), AI agents empower the interaction between humans and computers, therefore emerging as a system solution to place traffic prediction in the context of agent coordination. Many studies~\cite{sytlemeimwut2023,dengimwut2024} enhance collaborative interaction using AI agents, facilitate creative inspiration of users, and increase the comprehensibility of data. However, few studies tackled traffic-related areas with AI agents to increase the interactivity and comprehensibility of data. To better perform traffic flow analysis, transportation agents with generative AI paradigm show bright potentials to integrate user needs, traffic prediction, transportation management and mobile traveling as a holistic system for end-to-end spatial-temporal learning.

\vspace{1pt} \noindent \textbf{Our Approaches.} To address these issues, we proposed \textsf{TrafficGPT}, a multi-scale traffic generation system incorporated with a spatial-temporal agent framework. \textsf{TrafficGPT} contains three agents: a text-to-demand agent, a traffic prediction agent, and a suggestion and visualization agent. Firstly, in the text-to-demand agent, we used a pre-prompted Question \& Answer AI to extract users' demands, forming the specific traffic prediction tasks. Second, we proposed a transformer-based model to efficiently process multi-scale temporal data and limited spatial features, which enrich the representations of traffic conditions during traffic prediction. Accordingly, multi-scale analysis will be conducted and generate three types of prediction results. Last, based on the results, the suggestion and visualization agent will generate a series of interactive visualizations to promote users' comprehension of the prediction results. The proposed \textsf{TrafficGPT} system is depicted in Fig.~\ref{Abstract}.

\begin{figure*}[h]
    \centering
    \begin{tabular}{cc}
    \includegraphics[width=0.47\textwidth]{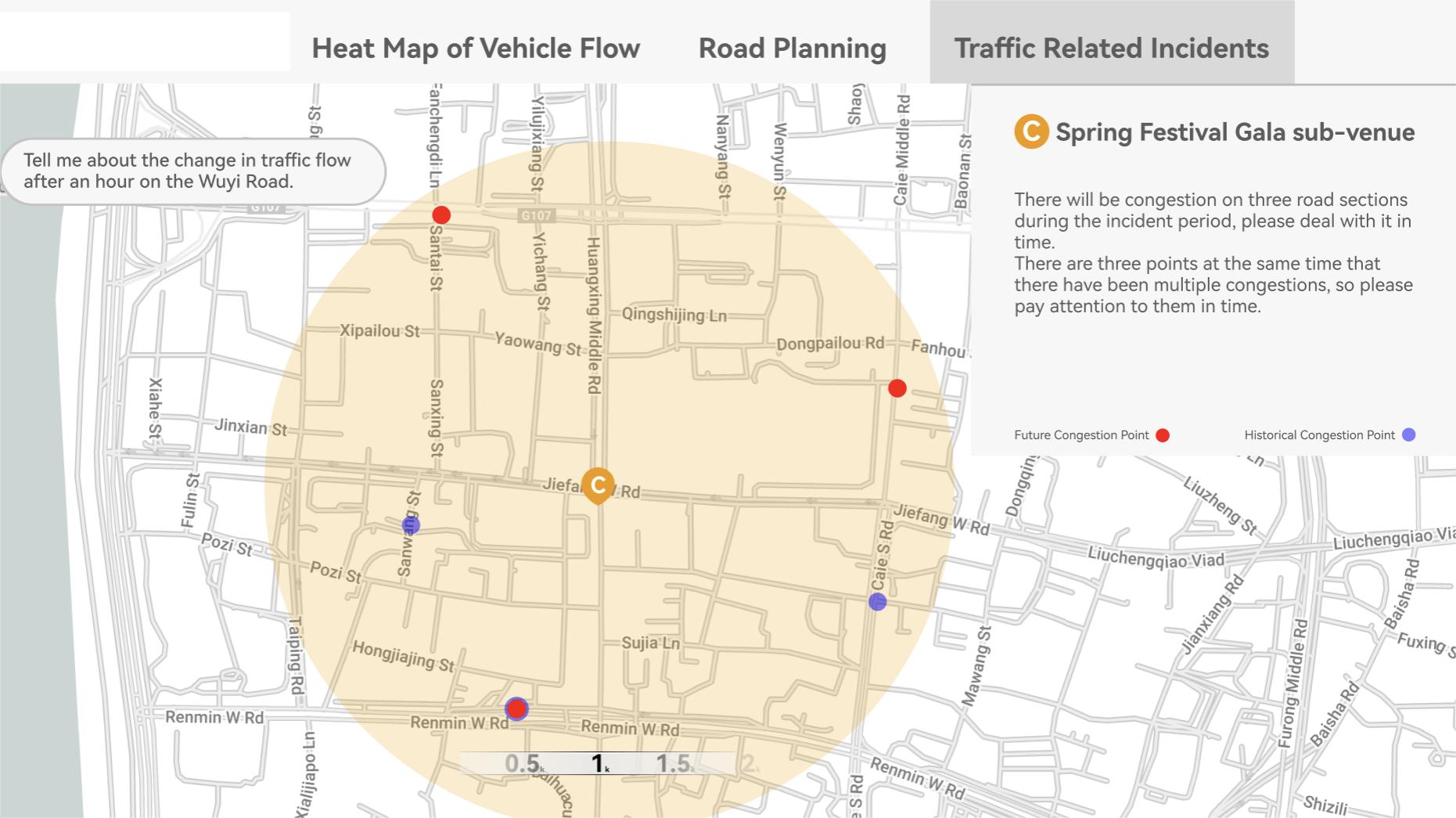}& \includegraphics[width=0.47\textwidth]{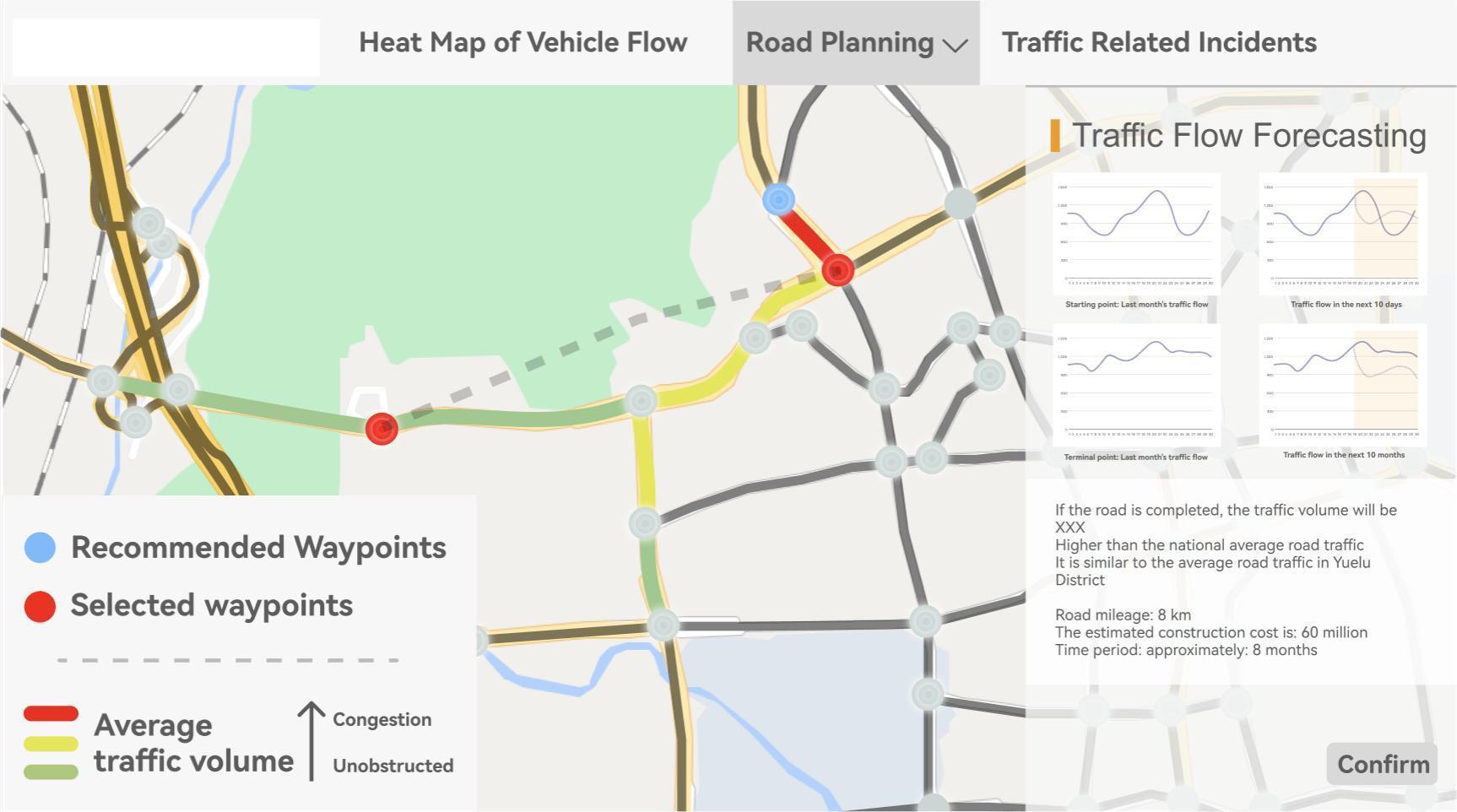} \\
    {\scriptsize (a) A user interacts with \textsf{TrafficGPT} using text.}&{\scriptsize (b) A user uses \textsf{TrafficGPT} to estimate an unseen road.}
     \end{tabular}
     \caption{The user interface of TrafficGPT system.}
    \label{Interface}
\end{figure*}

\vspace{1pt} \noindent \textbf{Contributions.} To the best of our knowledge, \textit{\textsf{TrafficGPT} is the first graph learning framework in transportation to address the coordination of different generative AI agents for end-to-end interaction of multi-scale traffic flow analysis from system perspective}. As illustrated by the interface of \textsf{TrafficGPT} in Fig.~\ref{Interface}, \textsf{TrafficGPT} is designed to solve various traffic-related ubiquitous computing concerns from transportation users. Our key contributions are listed as follows:
\begin{itemize}
\item \textbf{Dynamic Fusion for Multi-scale Transformer (DFMT).} We study the influence of periodic features and recent trend features on traffic data during the target period. The DMFT model designed for multi-scale temporal features is proposed to fuse the trend features of traffic data with periodic features, such as daily- and weekly-periodic. By balancing spatial and temporal dependencies, the DFMT model achieves comparable prediction accuracy with state-of-the-art models.
\item \textbf{Long-term prediction, unseen road estimation, and data analysis.} We devise training methods for long-term prediction and unseen road estimation, effectively using the semantic information shared by roads inside the road network. Detailed data analysis reveals the feasibility of rationally using spatial and temporal similarity to enhance traffic representations, extract co-semantic road network information, and rapidly adapt the model to new road networks.
\item \textbf{AI-agent system for traffic prediction.} We first propose a multi-scale traffic generation system, \textsf{TrafficGPT}, using spatial-temporal AI agents to promote the comprehensibility of traffic. We conduct comprehensive performance evaluations, including short-term prediction, long-term prediction, scalability, unseen road estimation, and interactive evaluation. We conduct the experiments on five real-world road traffic datasets (and one more dataset for extra experiments). The experimental results demonstrate that \textsf{TrafficGPT} has average SOTA performance on different prediction tasks and good interactive performance feedback from users.
\end{itemize}

\section{Related Work}
\label{Sec-Rel}
\subsection{Traffic Prediction using Neural Networks}
The most commonly used prediction methods for early studies in the field of traffic prediction are HA \cite{Wei2004ha}, ARIMA \cite{Han2004arima} and its variants, and Kalman filter \cite{Kalman2019} and other statistic-based schemes. However, because traffic data on arterial roads change nonlinearly, statistical techniques can hardly predict nonlinear traffic data \cite{2019Attention,Chen2020}, and thus the prediction accuracy may suffer significantly. Although KNN-based \cite{KNN2019} and SVR-based schemes \cite{2017ShortSVM,SVR2019} can model nonlinear traffic data, these machine learning-based schemes present limitations when capturing the temporal features of traffic patterns for arterial roads due to their lack of consideration of periodicity and time-varying between arterial roads \cite{temporal2020}. Faloutsos et al.~\cite{FaloutsosSIGMOD2019}, inspired the extraction of periodicity as an essential feature of time-series data.

Neural networks (e.g., RNNs \cite{ANN2018,SAE2018}) are also used to predict traffic data on arterial roads. However, RNNs are still plagued by gradient disappearance and gradient explosion. This problem can be mitigated by LSTM \cite{2015LSTM} and GRU \cite{GRU2018}. Although the RNN model is good at mining the temporal features of traffic data, it cannot capture spatial features. CNN \cite{2018CNN,2020CNN_LSTM,2020ConvLSTM} was introduced for traffic prediction to extract spatial relations of traffic data. Although great progress has been made in traffic prediction tasks by introducing CNN to mine the spatial dependence of traffic data, CNN is only suitable for mining Euclidean data, such as images and raster data \cite{2019Attention}. However, the traffic network does not accord with Euclidean data in essence and is a graph structure. With the development of the GCN model \cite{2020GCN,jia2020graphsleepnet,xu2019graph}, researchers use the model to mine the structural features of graph networks, providing a good solution to the above problems. Yu et al. \cite{YuIJCAI2018} propose a spatial-temporal Graph Convolutional Network (STGCN) to solve the time-series prediction problem in the traffic field. Zhao et al. \cite{2020GCN} propose a Temporal Graph Convolutional Network (T-GCN) based on GCN and GRU to predict traffic states. Wang et al.~\cite{WangSIGMOD2021} propose an asynchronous continuous-time dynamic graph algorithm for real-time temporal graph embedding. However, Zhao et al. and Wang et al. fail to fully extract temporal features. Although graph neural networks can be used to mine the spatial topology of traffic data well, capturing dynamic spatial-temporal relationships still requires further research.

\subsection{AI Agent with Human-AI Interaction}
The rapid development of artificial intelligence (AI) agents has facilitated the transition of human-computer interaction from the primary theoretical research stage to the practical application stage~\cite{wangimwut2020, diimwut2022}. This revolutionary advancement has facilitated the transportation field to evolve from a paradigm where humans use mathematical tools to analyze traffic to a collaborative framework where humans interact with traffic. 

Some studies have revealed the potential for using AI agents to optimize traffic policies. For example, Tang et al.~\cite{Yiqingdtpi2023} explored the catalytic role of ChatGPT for road policy improvement and they found that using ChatGPT for traffic control optimization can increase the average speed of vehicle travel. Villarreal et al.~\cite{Michaelitsc2023} discovered that novices can utilize reinforcement learning-related agents to solve complex hybrid flow control problems that typically require many areas of expertise to research. 
Meanwhile, some studies used AI agents to improve the interaction between humans and traffic. For instance, Amin et al.~\cite{Aminiui2024} found that machine learning algorithms greatly enhance data exploration in interactive visualizations, especially for users with limited domain knowledge. 

In the realm of traffic prediction, the biggest problem is the huge amount of data and the complex structure of spatial-temporal data, which makes it difficult to interpret in a generalized way for the transportation participants. Transportation participants aspire to understand the influence of prediction results, resulting in the challenge of interpreting the traffic prediction outputs. Directly visualizing them leads to the implicit assumption of unprofessional transportation participants. Therefore, as a general concern, increasing the interactivity between transportation participants and prediction results becomes influential in alleviating traffic-related concerns.





\section{Motivation and System Overview}
\label{Motivation}
Our motivation includes two parts: i) the accuracy improvement by multi-scale temporal data process;
ii) the system overview of TrafficGPT.

\subsection{Spatial-temporal Dependency}
To improve the performance of traffic prediction, some studies focus on using as much as spatial features, such as graph-based~\cite{wuijcai2019, songaaai2020} and recurrent-based~\cite{LIiclr2018, 2020TGCN} methods, which choose to separately process spatial features and temporal features. On the contrary, attention-based methods~\cite{xuarxiv2020, zhengaaai2020}, are adept at extracting the correlation of data and prefer to chemically fuse spatial and temporal features, by frequently using residual operation in their models. Compared with some graph-based methods and recurrent-based methods, attention-based methods have better performance. However, compared with spatial features, temporal features in these studies have not been effectively exploited.

Temporal information can be enriched by utilizing data from multiple scales of time. To confirm it, we have tested the short-term prediction performance on a dataset (Aliyun dataset\footnote{https://tianchi.aliyun.com/competition/entrance/231598/information?from=oldUrl}) and the results are presented in Table~\ref{Exp-mova}. The number of spatial-temporal blocks of STTN was set to 2 and we retained spatial components and temporal components of STTN as STTN\_S and STTN\_T, respectively. 
Then, we use multi-scale data as input of SNN, marked with $\*$. Three tasks of single-step and multi-step prediction were performed and used MAE, RMSE, and MAPE as evaluation metrics. Results indicate that spatial components are more effective than temporal components in STTN; meanwhile, multi-scale temporal extraction indeed enhances performance, particularly in terms of RMSE metrics.
\begin{table}[htb]
\small
\caption{Motivation experiment results.}
\renewcommand\arraystretch{1}
\centering
\label{Exp-mova}
\centering
\scalebox{0.7}[0.7]{$
\begin{tabular}{c|cccccccccc}
\hline
\multirow{2}{*}{Scheme} & \multicolumn{3}{c}{1 step prediction}  & \multicolumn{3}{c}{2 steps prediciton}  & \multicolumn{3}{c}{3 steps prediction}\\ \cline{2-10} 
& MAE  & RMSE & MAPE & MAE & RMSE & MAPE & MAE & RMSE & MAPE   \\ \hline
$STTN\_S$ & 3.28 & 8.12 & 19.58\% & 4.32 & 10.03 & 25.84\% &  4.94 & 11.29 & 24.79\% \\ 
$STTN\_T$ & 3.32 & 8.15 & 20.33\% & 4.70 & 10.71 & 28.96\% &  5.57 & 12.50 & 27.72\% \\
$STTN$ & 3.24 & 8.18 & 19.51\% & 4.29 & 10.16 & 25.14\% &  4.89 & 11.30 & 28.13\% \\ \hline
$STTN\_S^{\*}$ & 3.24 & 8.08 & 19.28\% & 4.30 & 9.99 & 25.21\% &  4.90 & 10.91 & 24.01\% \\
$STTN\_T^{\*}$ & 3.32 & 8.05 & 20.08\% & 4.67 & 10.63 & 28.76\% &  5.50 & 12.12 & 27.61\% \\
$STTN^{\*}$ & 3.22 & 8.12 & 19.51\% & 4.24 & 9.18 & 25.14\% &  4.80 & 10.70 & 24.26\% \\
\hline
\end{tabular}$}
\end{table}

\subsection{System Overview}
\begin{figure*}[tb]
    \centering
    \includegraphics[width=0.8\textwidth]{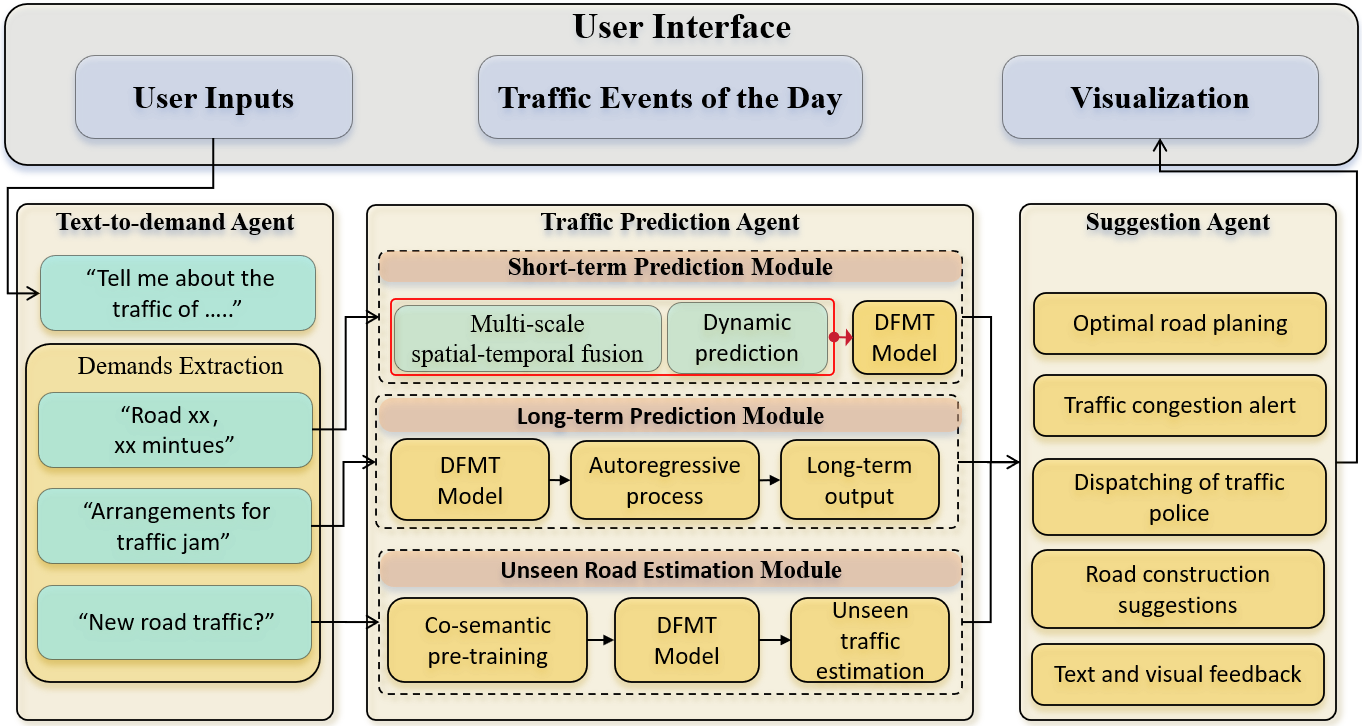}
    \caption{An overview of our TrafficGPT system.}
    \label{System}
\end{figure*}

The system framework of \textsf{TrafficGPT} is depicted in Fig.~\ref{System}. On the upper side (gray) is the user interface, which contains the functions of collecting users' input text, presenting traffic events, and visualizing the prediction results. On the lower side, the system comprises three main components: The Text-to-demand Agent, extracts the prediction requirements from the input text and activates corresponding modules in the next agent. The Traffic Prediction Agent consists of three modules: short-term prediction, long-term prediction, and unseen road estimation, each of which corresponds to different traffic prediction tasks. The Suggestion Agent is comprised of functions to maximize the use of prediction results to provide appropriate traffic suggestions; comprehensible visualizations will be provided based on the prediction results and generated suggestions.

\begin{itemize}
    \item \textbf{Text-to-demand Agent.} Users with limited domain knowledge in traffic prediction, often fail to precisely describe and realize their requirements for predicting traffic conditions. This action agnosticism can cause users to grow anxious about traffic. Our method deploys a pre-prompted Q\&A AI agent based on GPT~\cite{GPT3} to precisely extract the specific prediction tasks from users. Users can provide text to describe their traffic needs and the agent automatically extracts corresponding tasks and conveys them to the next agent.
    
    \item \textbf{Traffic Prediction Agent.} In the traffic prediction agent, three modules for different prediction modules are activated by the conveyed tasks, and use multi-scale data to predict. The short-term prediction module, enable accurate prediction for 1-hour traffic in the future by using the proposed Dynamic Fusion Multi-scale Transformer (DMFT) model. The long-term prediction module uses an autoregressive algorithm to facilitate the DMFT model with the perception of long-term (a few days, months, etc.) traffic, allowing the DMFT model to generate precise long-term prediction results. The unseen road estimation module use two-stage training method to extract co-semantic information by selecting highly relevant roads using spatial-temporal similarity. These modules maximize the utilization of the multi-scale data to enhance its interpretability for users.
    
    \item \textbf{Suggestion Agent.} To enhance the comprehensibility and interactivity of our system, the suggestion agent receives the prediction results and generates multiple suggestions for users to directly understand the traffic. Based on some existing algorithms, we can enrich this agent with more functions by rationally using three types of prediction results and their combinations. This agent increases non-professional users' perception of traffic, ensuring effective interactions between users and traffic data. Additionally, diverse visualizations based on the results and suggestions are generated to empower the system with high comprehensibility, consequently alleviating users traffic-related anxiety.
    
\end{itemize}

\section{TrafficGPT Design}
\label{Sec-Model}
The design of TrafficGPT centers on using multi-scale data to conduct multi-scale traffic analysis and realizing the multi-scale visualization for user-computer interaction. In this section, we focus on introducing an innovative traffic prediction agent, consisting of a short-term prediction module, a long-term prediction module, and an unseen road estimation module. Then, we use some existing and successful algorithms to build up the text-to-demand agent and suggestion agent for visualizations. 

\subsection{Problem Formulation}

\label{Sec-PD}
Problem formulation for traffic state forecasting is set as the priority to clarify the problems to be solved in our system. Traffic prediction involves predicting future traffic information for a given period based on the road's historical traffic data. Notably, the traffic information mentioned here is a general concept \cite{2020TGCN}. This prediction problem is affected by both upstream and downstream historical traffic information and is influenced by similar traffic patterns in distant regions. 

To address these problems, we propose a Dynamic Fusion Multi-scale Transformer (DFMT) model. 
Specifically, we use two graph structures to represent the spatial relations between different roads in the traffic network. The connectivity graph $G_r$, an unweighted graph, describes the topological structure of the road network. The correlation graph $G_c$, a weighted graph, represents the correlations between different roads and assigns greater weights to roads with similar traffic patterns.

\begin{figure*}[htb]
    \centering
    \includegraphics[width=0.95\textwidth]{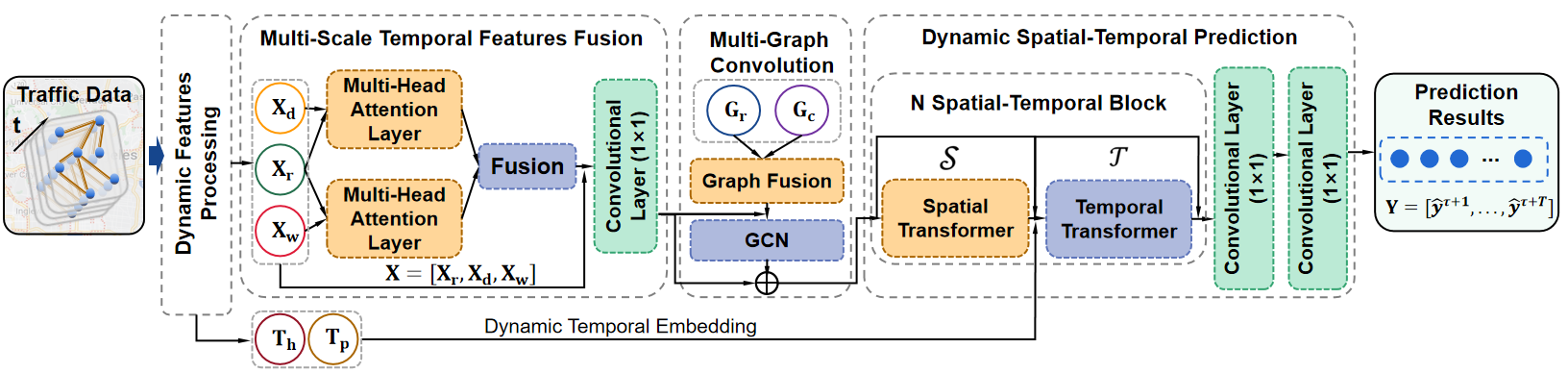}
    \caption{The structure of the proposed Dynamic Fusion Multi-scale Transformer (DFMT) model.}
    \label{framework}
\end{figure*}

\vspace{3pt} \noindent \textbf{Structure of framework.} 
To elaborate, the structure of framework (see Fig.~\ref{framework}) is divided into three major layers:
(1) Layer 1 (multi-scale temporal feature fusion) aims to capture the temporal features (e.g. daily and weekly periodicity) of the traffic data.
(2) Layer 2 (multi-graph convolution and spatial transformer) is to extract the spatial features (e.g. topological structure and traffic correlation) of the traffic data.
(3) Layer 3 (dynamic spatial-temporal prediction) is designed to tradeoff, integrate, and learn the fused spatial and temporal features from Layer 1 and Layer 2.
\vspace{3pt} \noindent \textbf{Connectivity graph $G_r$.} 
We denote by $G_r=(V_r,E_r,A_r)$ an undirected connected graph, which represents a road network topology. Each node in $G_r$ corresponds to a road. Let $V_r = \{v_1, \dots, v_i, \dots, v_N\}$ be the set of nodes in $G_r$, where $N$ is the number of roads. Let $v_i$ represent the $i$-th node of $V_r$, which represents the $i$-th road of the road network $G_r$. Let $ E_r = \{e_1, \dots, e_l, \dots, e_m\}$ be the edge set of $G_r$, where $m$ denotes the number of edges. Let $e_l = (v_i, v_j)$ be the edge between the $i$-th and $j$-th nodes of $G_r$, indicating that the $i$-th and $j$-th roads are adjacent. The edges here have no direction. Let ${A_r}\in R^{N\times N}$ be the adjacency matrix of $G_r$. Each element of $A_r$ is a binary value of 0 or 1, indicating connectivity between two roads. For example, $A_r[i][j] = A_r[j][i] = 1$ indicates that there is an edge between the $i$-th and $j$-th roads.

\vspace{3pt} \noindent \textbf{Correlation graph $G_c$.} 
With arterial roads' historical traffic information, we also build the correlation graph to measure the correlation of traffic information between different arterial roads. Let $G_c=(V_c,E_c,A_c)$ be an undirected, weighted, and connected graph, which represents the correlation of arterial roads' historical traffic information. Let $V_c$ be the node set of $G_c$. Let $E_c$ be the edge set of $G_c$. $A_c$ denotes the weighted matrix of the correlation graph $G_c$. In this paper, we use the Pearson correlation coefficient to calculate the correlation between historical traffic information for different arterial roads. Let $r_{i, j}$ be the Pearson correlation coefficient between arterial roads $i$ and $j$. The weight matrix $A_c$ of the correlation graph $G_c$ can be represented as follows:

\begin{equation}
r_{ij}=\frac{\sum_{k=1}^P ({X_k}^i - \bar{X^i})({X_k}^j - \bar{X^j})}{\sqrt{\sum_{k=1}^P {({X_k}^i - \bar{X^i})}^2} \sqrt{\sum_{k=1}^P {({X_k}^j - \bar{X^j})}^2}},
\end{equation}


\begin{equation}
{A_{c}}_{\{n,p\}}= r_{n p},\ n\in [0,N), p \in [0,P)
\end{equation}
where $N$ represents the number of nodes, and $P$ represents the number of samples (length of the historical time series). It is worth noting that the correlation graph is calculated by only training data.

\vspace{3pt} \noindent \textbf{Feature matrix $X^{N \times P}$.} 
The original data can be defined as $X=(X_r, X_d, X_w)$, which is regarded as historical traffic data. Let $X_r=(X_r^1, X_r^2, \dots, X_r^T) \in R^{N \times T}$ be a historical traffic data of all arterial roads in the $T$ time slices, which denotes a segment of historical traffic data directly adjacent to the prediction period. Let $X_r^i$ be the recent traffic data of the $t_i$-th prediction period, where $T_r$ denotes the length of time slices in the recent trend. Here, $X_r^i = (X_{(t_i-T_r)}, X_{(t_i-T_r+1)}, X_{(t_i-T_r+2)}, \dots, X_{(t_i-1)}) \in R^{N \times T_r } $.  Let $X_d=(X_d^1, X_d^2, \dots, X_d^T) \in R^{N \times T}$ be the daily-periodic traffic data of the prediction period. Let $ X_d^i$ be the daily-periodic traffic data of the $t_i$-th prediction period. 

Here, $ X_d^i = (X_{(t_i-T_d \times q)}, X_{(t_i-(T_d-1) \times q)}, \dots, X_{(t_i-q)}) \in R^{N \times T_d } $, where  $T_d$ denotes the length of time slices of the daily-periodic, and $q$ denotes the sampling frequency per day. Let $X_w=(X_w^1, X_w^2, \dots, X_w^T) \in R^{N \times T}$ be the weekly-periodic traffic data of the prediction period. Meanwhile, we set $ X_w^i$ as the weekly-periodic traffic data of the $t_i$-th prediction period $ X_w^i = (X_{(t_i-7 \times T_w \times q)},\ X_{(t_i-7 \times (T_w-1) \times q)}, \dots, X_{(t_i- 7 \times q)}) \in R^{N \times T_w } $, where $T_w$ denotes the length of time slices of the weekly-periodic. 

Thus, based on the topology information of road networks (i.e., $G_r$), the correlation of traffic information between different arterial roads (i.e., $G_c$), and the historical traffic information (i.e., $X$), the problem we study is to predict the future traffic information in the next $\tau$ periods by learning a mapping function $f$, as shown in Eq.~\ref{YC}:
\begin{equation}
\label{YC}
  [Y_{t+1}, \dots, Y_{t+\tau}]=f(G_r; G_c; X),
\end{equation}
where $[Y_{t+1}, \dots, Y_{t+\tau}]$ is future traffic data in the next $\tau$ periods.

In the temporal dimension, there are tendencies, daily periodicity, and weekly periodicity in traffic data. We utilize two identical multi-head self-attention mechanisms to fuse the recently-periodic (i.e., $X_r$), daily-periodic (i.e., $X_d$), and weekly-periodic (i.e., $X_w$) features of traffic data for mining the temporal features.

\begin{figure*}[tb]
    \begin{minipage}{0.45\textwidth}
        \centering
        \includegraphics[width=0.9\linewidth]{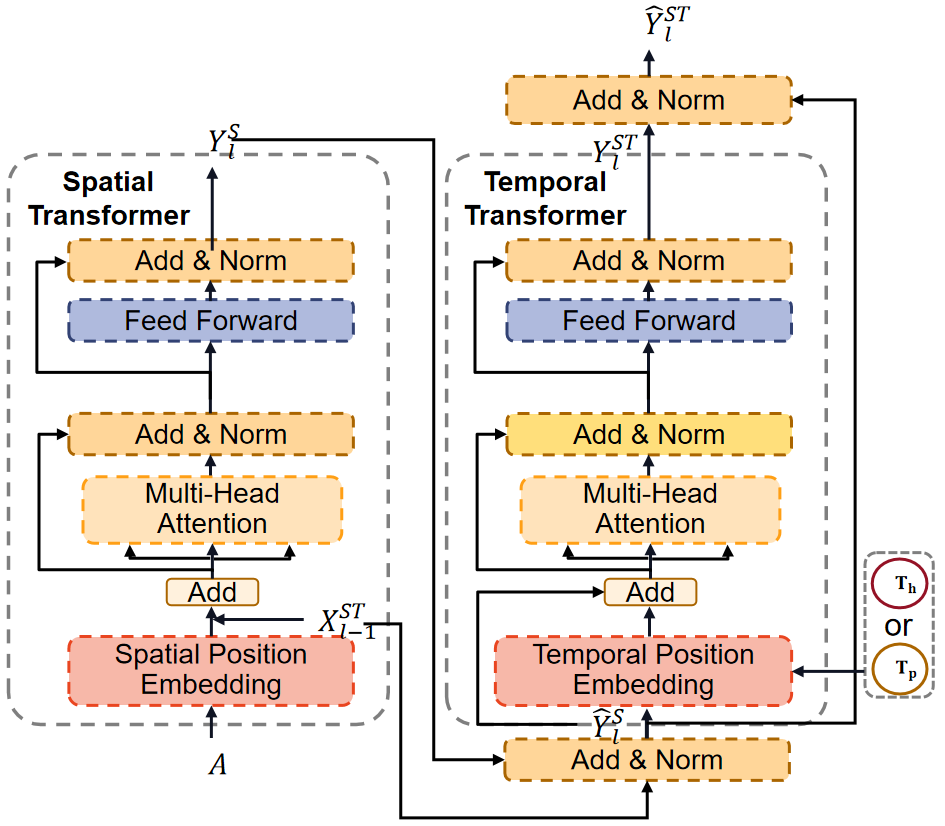}
        \caption{The structure of single spatial-temporal block.}
        \label{st_block}
    \end{minipage}%
    \begin{minipage}{0.55\textwidth}
        \centering
        \includegraphics[width=\linewidth]{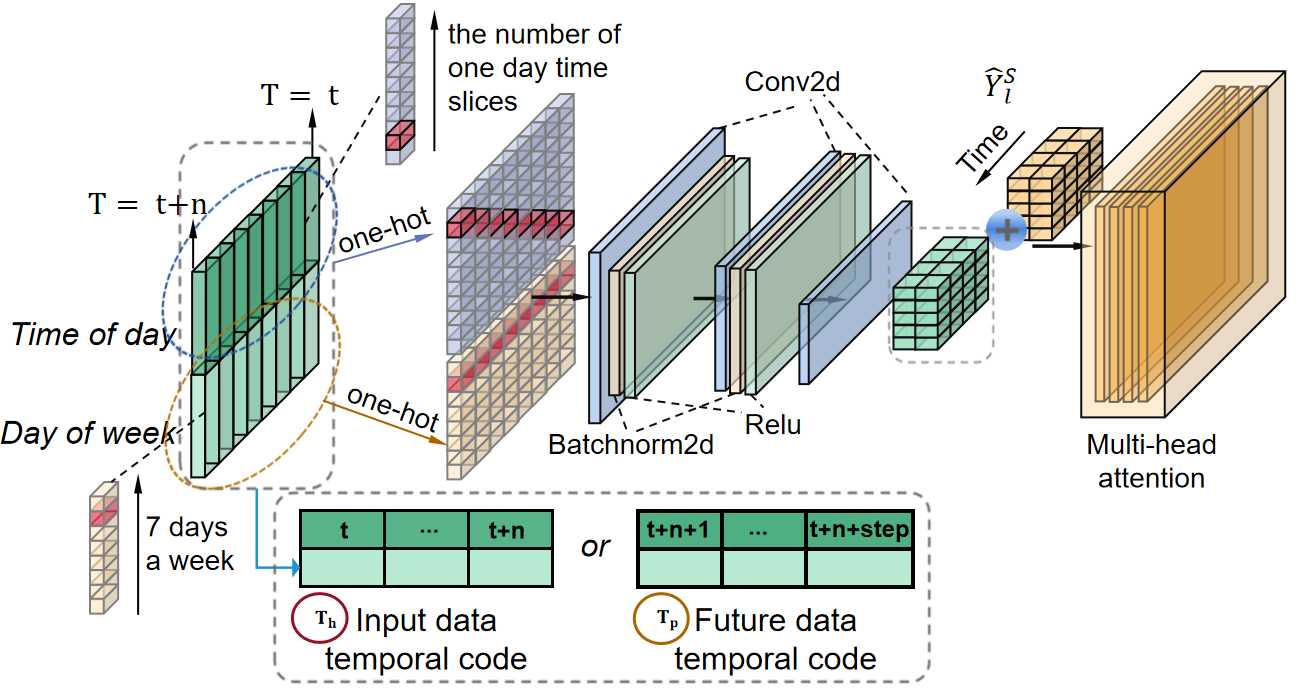}
        \caption{The detailed procedure of temporal embedding.}
        \label{fig:temporal_embed}
    \end{minipage}
\end{figure*}

\subsection{Short-term Prediction Module Design}
Multi-scale information of traffic data has strong potential for enhancing the quality of data extraction and thus increased accuracy. We proposed a DFMT model for effectively utilizing the information, to ensure short-term prediction accuracy. The components of DFMT are presented in Fig.~\ref{framework}, i.e., the multi-scale temporal feature fusion module, the multi-graph convolution module, and the dynamic spatial-temporal prediction module.

\subsubsection{Multi-Scale Temporal Feature Fusion}
The initial inputs $X_r$, $X_d$, and $X_w$ are transformed to $X_{rd}=(X_r, X_d)$ and $X_{rw}=(X_r, X_w)$, as the feeds of multi-head attention layers depicted in Fig.~\ref{framework}. Subsequently, the same treatment is done for both multi-head attention layers. The calculation formulas of both multi-head attention layers, which differ from the first multi-head attention mechanism only in the data input, are
\begin{equation}
      \label{Eq.aw}
      \scalebox{0.9}{$
	 {\alpha}^{j}_{rw}= {{Q}^{j}_{rw} {K^{j}_{rw}}^{\top}} /{\sqrt{d^{j}_{q_{rw},k_{rw}}}} ;\
	 \hat{{\alpha}}^{j}_{rw}= \text{Softmax}({\alpha}^{j}_{rw})
  $}
\end{equation}
where ${Q}^{j}_{rw}= W^{Q^{j}_{rw}} X_{rw}$, ${K}^{j}_{rw}= W^{K^{j}_{rw}} X_{rw}$, and ${V}^{j}_{rw}= W^{V^{j}_{rw}} X_{rw}$ denote the query subspaces of the $j$-th head of multi-head attention mechanism, respectively. Here, $W^{Q^{j}_{rw}}$, $W^{K^{j}_{rw}}$, and $W^{V^{j}_{rw}}$ are the weight matrices for ${Q}^{j}_{rw}$, ${K}^{j}_{rw}$, and ${V}^{j}_{rw}$, respectively. ${M}^{j}_{rw}= \hat{{\alpha}}^{j}_{rw} {V}^{j}_{rw}$ is set as the output of the $j$-th self-attention layer of  multi-head attention mechanism to compute the ${\hat{M}}_{rw}={Concat}({M}^{1}_{rw}, \dots, {M}^{j}_{rw}, \dots, {M}^{h}_{rw})$. Then, the output of multi-head attention mechanism, ${M}_{rw}$, would then be computed from the formula ${M}_{rw}={\hat{M}}_{rw} W_{rw}$. 

Then, to obtain the multi-scale temporal feature fusion result, we fuse the results of the two multi-head attention mechanisms and make the residual connection with the feature matrix $X$. Finally, we use a single $1 \times 1$ convolutional layer to aggregate these temporal features as input to the following multi-graph convolution module. The formula for multi-scale temporal feature fusion is as follows:
\begin{equation}
      \label{Eq.fusion}
	 X^T= {Conv} ( W^{'}_{rd} M_{rd} + W^{'}_{rw} M_{rw} + X ),
\end{equation}
where $X^T$ denotes the temporal features after multi-scale temporal feature fusion.

\subsubsection{Multi-Graph Convolution}


As mentioned, we construct two graphs $G_r$ and $G_c$, which have been defined in the previous Section Problem Formulation, for the road network to reflect the spatial relations between heterogeneous arterial roads. $G_r$ and $G_c$ will be integrated in the multi-graph convolution module. The multi-graph convolution module mainly consists of two parts: Graph Fusion and Graph Convolution Networks. 

\vspace{3pt} \noindent \textbf{Graph Fusion} is to fuse the adjacency matrix of the connectivity graph and the weight matrix of the correlation graph into the final graph weight matrix. We first standardize the adjacency matrix of the connectivity graph and the weight matrix of the correlation graph, using the formulas below:
\begin{equation}
      \label{Eq.standardize}
	 \hat A_r= \hat D_r^{- \frac {1}{2}} (A_r + I) \hat D_r^{- \frac {1}{2}} ;\ \hat A_c= \hat D_c^{- \frac {1}{2}} (A_c + I) \hat D_c^{- \frac {1}{2}}
\end{equation}

Then, $\hat A_r$ and $\hat A_c$ are fused into a final graph weight matrix by element-level weighted summation. The formulas of the graph fusion process are as follows:
\begin{equation}
	 W^{'}_r, W^{'}_c = {Softmax}(W_r, W_c) ;\ A =  W^{'}_r \hat A_r + W^{'}_c \hat A_c
\end{equation}
where $A$ denotes the multi-graph fusion weight matrix, $W^{'}_r $ and $W^{'}_c$ are the weight matrices of $\hat A_r$ and $\hat A_c$, respectively. The softmax operation can maintain the result of the weighted sum operation normalized. 

\vspace{3pt} \noindent \textbf{Graph Convolution Networks} mainly use the multi-scale temporal features fusion results and the multi-graph fusion weight matrix to carry out graph convolution operations and residual connection. We adopt two-layer graph convolution to extract spatial features of traffic data. The formula of the graph convolution process is $X^{ST}= A W_2 ( A W_1 X^T) + X^T$,
where $W_1$ and $W_2$ are the convolution weight matrices, and $X^{ST}$ is the multi-graph convolution result which denotes the spatial-temporal features of arterial roads. We further mine the spatial features of traffic data through the multi-graph convolution module, and then the spatial-temporal features $X^{ST}$ and the multi-graph fusion weight matrix $A$ are used as the input of the next dynamic spatial-temporal prediction module.
\begin{figure*}[ht]
    \centering
    \includegraphics[width=0.95\textwidth]{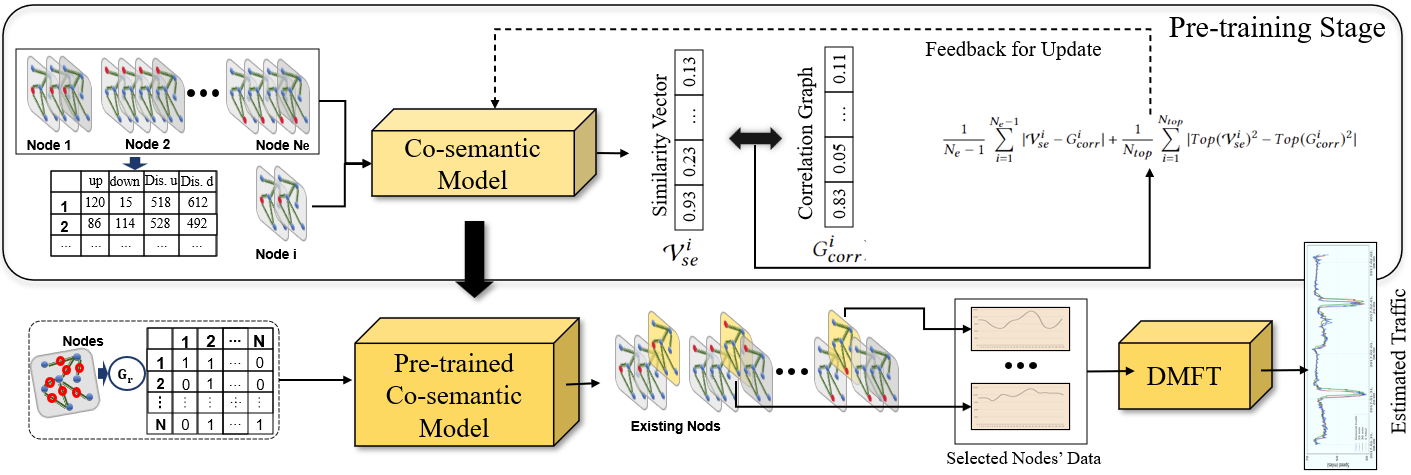}
    \caption{Two stages of unseen road estimation training.}
    \label{fig:estimate}
\end{figure*}
\subsubsection{Dynamic Spatial-Temporal Prediction}

The dynamic spatial-temporal prediction module utilizes $N$ spatial-temporal blocks and two $1\times 1$ convolution layers to predict traffic data. Each spatial-temporal block consists of a spatial transformer and a temporal transformer in series, and they jointly mine spatial-temporal features of traffic data. The structure of a single spatial-temporal block is shown in Fig.~\ref{st_block}. 
In the spatial-temporal block, the spatial transformer $\mathcal{S}$ mines dynamic spatial features of arterial roads from the spatial-temporal features $X^{ST}$ and the multi-graph fusion weight matrix $A$, the temporal transformer $\mathcal{T}$ further extracts dynamic temporal features of arterial roads from the spatial-temporal features $X^{ST}$ and the output results of the spatial transformer $\mathcal{S}$. Residual connection and layer normalization are adopted to stabilize training after the spatial transformer and the temporal transformer of each spatial-temporal block. For extracting dynamic spatial-temporal features from the $l$ spatial-temporal block, we firstly process the input $X^{ST}_{l-1}$ by $Y^{S}_{l}=\mathcal{S} ( X^{ST}_{l-1}, A )$ to obtain the $Y^{S}_{l}$, which is the output of the spatial transformer of the $l$-th spatial-temporal block. Then, more procedures are introduced to extraction: $\hat Y^{S}_{l}={LayerNorm} (Y^{S}_{l} + X^{ST}_{l-1})$, $Y^{ST}_{l}=\mathcal{T} ( \hat Y^{S}_{l} )$, and $\hat Y^{ST}_{l}={LayerNorm} (Y^{ST}_{l} + \hat Y^{S}_{l})$ 
where $\hat Y^{S}_{l}$ represents the result of the spatial transformer of the $l$-th spatial-temporal block after residual connection and layer normalization. $Y^{ST}_{l}$ denotes the output of the temporal transformer of the $l$-th spatial-temporal block. $\hat Y^{ST}_{l}$ represents the output result of the $l$-th spatial-temporal block. And then we take $X^{ST}_{l}= \hat Y^{ST}_{l} $ as the input for the next spatial-temporal block.
The detailed temporal embedding procedure is illustrated in Fig.~\ref{fig:temporal_embed}. It is worth noticing that $T_p$ as the future temporal code would be only used in the last spatial-temporal block. $T_h$ is used in other blocks. This process provides DFMT a perspective to understand the temporal relationship between the input data and predict data. One hot encoding mechanism and three convolution layers are involved to refine the temporal features inside the $T_h$ or $T_p$. After concentrating the $T_h$ or $T_p$ with $\hat Y^{S}_{l}$ in time dimension, the mix features would be subsequently fed into the multi-head attention module.

After $N$ spatial-temporal blocks, two $1 \times 1$ convolution layers are used to implement the prediction. The first convolution layer's input is the feature output from the previous spatial-temporal block $Y^{ST}_N$. The multi-step prediction results $[Y_{t+1}, \dots, Y_{t+\tau}]$ in future $\tau$ periods are calculated as $[Y_{t+1}, \dots, Y_{t+\tau}] = {Conv}( {Relu} ({Conv}(Y^{ST}_N) ) ).$


\subsection{Long-term Prediction Module Design}
\label{sec:long-term method}
\subsubsection{Autoregressive Process}
The autoregressive process uses the first stage output as the next stage input. Assuming the DFMT model is well-trained in the short-term module and the road network is unchanged, we use the autoregressive process to fine-tune the model. We randomly select a period of data $D^s$ to fine-tune. Specifically, graphs of $G_r$ and $G_c$ will be computed as the same as Section~\ref{Sec-PD}, and we use the first 12 steps of the data as the only input feature matrix $X^f=(X_r^f, X_d^f, X_w^f)$. Set the real temporal features of this slice start at $t^r$, the $X_r^f$ will equal to $(X_{(t^r-12)}, X_{(t^r-11)}, X_{(t^r-10)}, \dots, X_{(t^r-1)}) \in R^{N \times 12}$.
Additionally, $X_d^f, X_w^f$ equal to $(X_{(t^r-T_d \times q)}, X_{(t^r-(T_d-1) \times q)}, \dots, X_{(t^r-q)}) \in R^{N \times T_d } $ and $ (X_{(t^r-7 \times T_w \times q)},\ X_{(t^r-7 \times (T_w-1) \times q)}, \dots, X_{(t^r- 7 \times q)}) \in R^{N \times T_w }$. After the first stage training, the first stage output is $Y^r_1 = [Y_{t^r+1}, \dots, Y_{t^r+12}]$. In the second stage, $Y^r_1$ will be input in the model and generate the next stage output $Y^r_2$. After generating $Y_{L_g}^r$, where $L_g = Length(D^s) -12$, we constructing a $Y_{L_g}^p$ by collecting the first value of $Y^r_i$, where $i=1, 2, ..., L_g$. Then, we compute the $L1loss$ with the selected period of data $D^s$ (without the first 12 steps) to fine-tune the model. The process can be summarized as follows:

\begin{equation}
        input_i =
    \begin{cases}
    X^f, & \text{if }  i = 1 \\
     Y_{i-1}, & \text{if } 1 < i \leq L_g
    \end{cases}
\end{equation}

\subsubsection{Long-term Output Module}
After the fine-tuning process, the model can use the first stage input $X^f$  to produce the first stage output $Y^r_1$ and use $Y^r_2$. During the autoregressive testing, the short range of the ahead prediction suffers accuracy fluctuation, however, accuracy will be stable soon. Specifically, the first-day prediction may not be as confident as the prediction after the first day. This phenomenon will be present in Section~\ref{sec:long-term:exp}, where the results indicate that this method is more suitable for long-period traffic prediction, which is stable and accurate.

\subsection{Unseen Road Estimation Module Design}
\label{sec:unseen pre}
Traffic on different roads in the same road network shares the same semantic information. We proposed a co-semantic extraction method to estimate the traffic of unseen roads that will be added to the existing road network. However, not all nodes can contribute to accurate estimating. To eliminate the interference of unrelated nodes, our method will select the most relevant nodes based on spatial and temporal similarity. The selected nodes should interact with the road network in a similar way (number of connections, distance between connections) and have high traffic correlation.

\subsubsection{Problem Formulation}
We set that the existing road network has $N_e$ nodes; the spatial information of the road network contains the connectivity of nodes and the distance between each pair of connected nodes. Meanwhile, we use historical data from existing nodes to generate a correlation graph, $G_{corr}$. 
Subsequently, $N_{un}$ of unseen nodes, which will connect to the existing road network in the future, are proposed for training and testing.

Each node (existing or unseen) has a corresponding spatial semantic graph $G_{ss} = (C_{up}, C_{down}, D_{up}, D_{down})$, where $C_{up}$ denotes its up-stream nodes, $ C_{down}$ denotes its down-stream nodes, $D_{up}$ denotes the distance to up-stream nodes, and $D_{down}$ denotes the distance to down-stream nodes. We set the spatial semantic graph of the existing nodes and unseen nodes as $G_{se} = (G_{se}^1, G_{se}^2,..., G_{se}^i)$ and $G_{su}= (G_{su}^{1}, G_{su}^{2},..., G_{su}^{j})$ respectively. Here, $i = 1, 2..., N_e$ and $j = 1, 2,..., N_{un}$.

\subsubsection{Co-semantic Extraction and Traffic Estimation}
To ensure the generalization ability of the estimation, we designed a co-semantic model based on spatial similarity and traffic correlation to select nodes. Two stages are included: the first stage is for pre-training the co-semantic model, and the second stage will train the DFMT model combined with the pre-trained co-semantic model. 

\vspace{3pt} \noindent \textbf{Pre-training the co-semantic Model.} The pre-training process used existing nodes to implement. We use $G_{se}$ as pre-training data and each node $G_{se}^i$ will be calculated the similarity with other node:
\begin{equation}
\scriptsize
\begin{split}
    \mathcal{V}^{i}_{se} = [Sim(G_{se}^1, G_{se}^i), ..., Sim(G_{se}^{i-1}, G_{se}^i), \\Sim(G_{se}^{i+1}, G_{se}^i), ..., Sim(G_{se}^{N_{e}}, G_{se}^i)]
\end{split}
\end{equation}
where $\mathcal{V}^{i}_{se}$ represents the similarity vector of $G_{se}^i$, and $Sim(*)$ represents the similarity function based on cosine similarity. Specifically, spatial graphs will be mapped to 1024 latent space. 

$Sim(G_{se}^1, G_{se}^i) = Cosine\_sim(Linear(G_{se}^1), Linear(G_{se}^I))$ is cosine similarit. Then we calculate the loss between $\mathcal{V}^{i}_{se}$ and $G_{corr}$ to update the model:
\begin{equation}
\scriptsize
\label{eq:loss_pre}
    Loss_p = \frac{1}{N_{e}-1} \sum_{i=1}^{N_{e}-1} |\mathcal{V}^{i}_{se} - G_{corr}^i| + \frac{1}{N_{top}} \sum_{i=1}^{N_{top}} |Top(\mathcal{V}^{i}_{se})^2 - Top(G_{corr}^i)^2|
\end{equation}
where $Loss_p$ denotes pre-training loss, $Top(*)$ denotes the function to find the top $N_{top}$ values, and the second term of Eq.~\ref{eq:loss_pre} is to amplify the impacts of traffic correlation between nodes. After pre-training, the co-semantic model can precisely select useful nodes for estimation.

\vspace{3pt} \noindent \textbf{Estimation Training.} In the second stage, we orderly input $G_{su}^j$ into the co-semantic model to get $\mathcal{V}^{i}_{su}$, representing the similarity vector between the unseen node and existing nodes. Subsequently, the top $k$ of existing nodes will be selected, where we set $k = 10$. Accordingly, we generate the corresponding connectivity graph $G_r^L$ and correlation graph $G_c^L$ (computed with selected nodes and all existing nodes), as mentioned in Section~\ref{Sec-PD}. Therefore, the $G_r^L$ and $G_c^L$ are not square matrices and will affect the generation of the degree matrix in the Laplace matrix computation. To solve it, we use an MLP model to map the graphs into square matrices. Moreover, the initialized parameters and updated parameters of the MLP model must be positive. 
Subsequently, we use the corresponding nodes' historical data $X^f$ as input to train the entire model and output the estimated data of the unseen node. Here, $X^L=(X_r^L, X_d^L, X_w^L) \in R^{3 \times k \times T_z}$ and $T_Z = T_r, T_d, or T_w$. The entire training process is depicted in Fig.~\ref{fig:estimate}.

\begin{figure*}[htb]
	\begin{center}
		\begin{tabular}{ccc}
			\includegraphics[width=0.6\columnwidth]{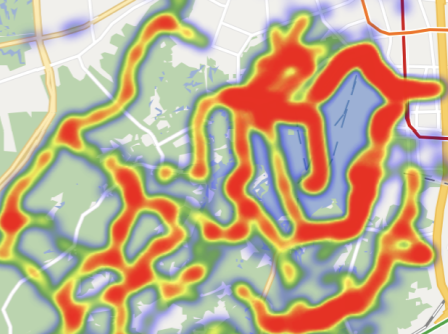}&
			\includegraphics[width=0.6\columnwidth]{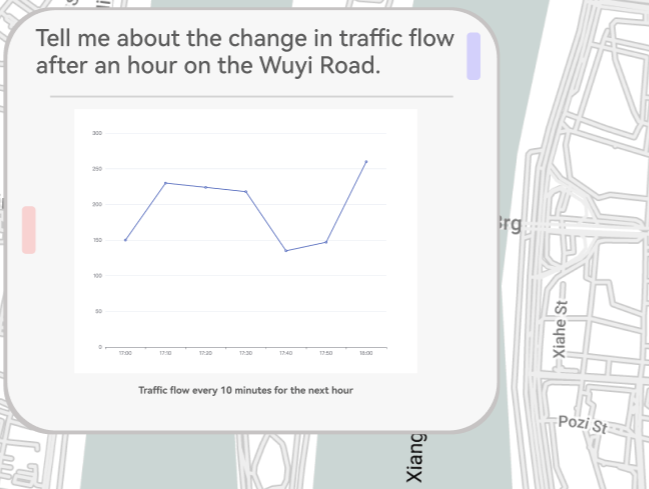}&
            \includegraphics[width=0.6\columnwidth]{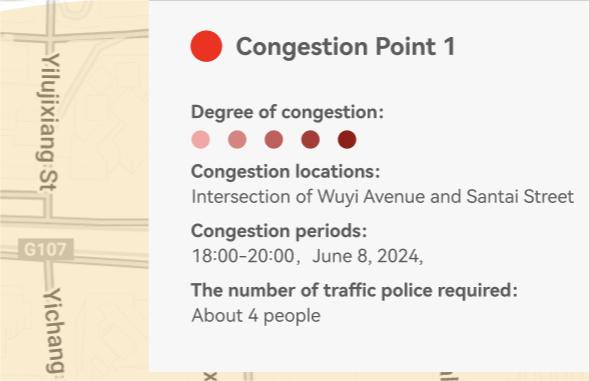}\\{\scriptsize (a) Heat Map.}& {\scriptsize (b) Text-to-demands.}& {\scriptsize (c) Congestion alert.} \\ 
            \includegraphics[width=0.6\columnwidth]{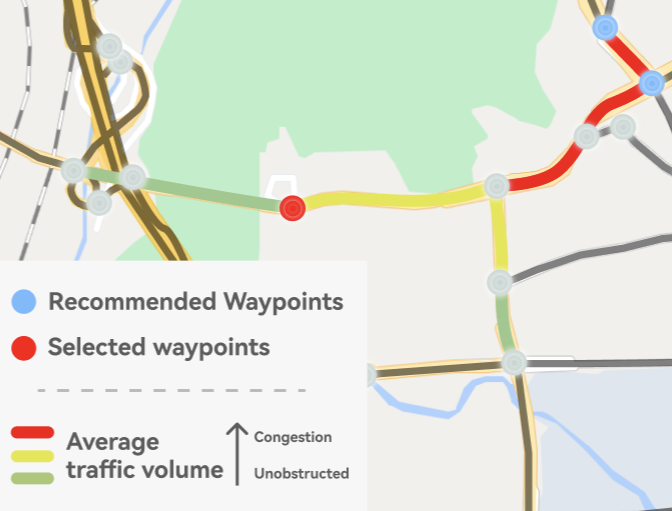} & \includegraphics[width=0.6\columnwidth]{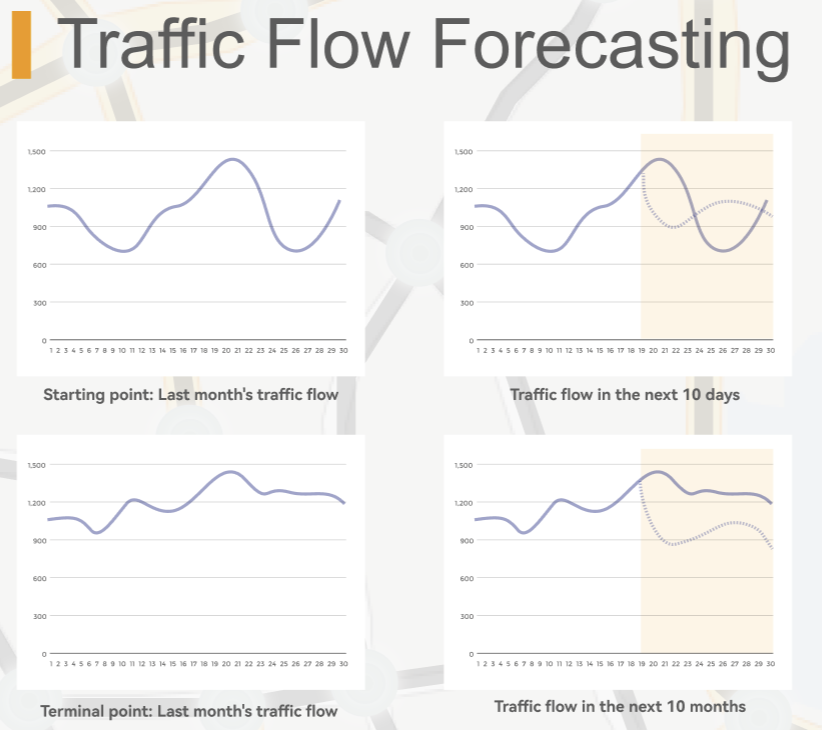}&
            \includegraphics[width=0.6\columnwidth]{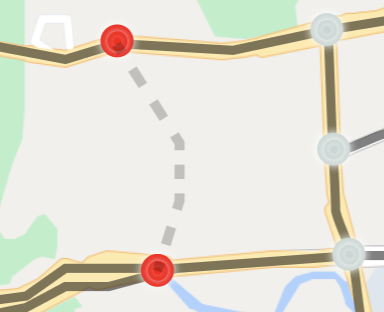}\\
			{\scriptsize (d) Optimal route planning.} &    {\scriptsize (e) Long-term future traffic.} & {\scriptsize (f) Unseen road estimation.} 
		\end{tabular}
	\end{center}
	\caption{Functions and visualizations.}
	\label{fig:vis_agent}
\end{figure*}

\subsection{Text-to-demmand Agent and Suggestion Agent}

AI empowers computers with the ability to understand human requirements. Many AI Q\&A-related works have been very successful, e.g., GPT3~\cite{GPT3}/GPT4~\cite{gpt4}, and Q\&A interactions are essential for nowadays AI agent systems. Meanwhile, many transportation suggestion-making algorithms are relatively mature and can be directly employed in our system. For example, using the A-star~\cite{kabir2022somet} and Dijkstra~\cite{zhou2023CNIOT} algorithm for optimal path planning. We used some existing algorithms to build and validate our system. 

\vspace{3pt} \noindent \textbf{Demands Extraction.} To facilitate our system with demand extraction ability, we used the open-source interface of GPT3~\cite{GPT3} and pre-prompt it into a traffic prediction scenario. We have used three questions to pre-prompt the GPT: (1) "Now I am a transportation participant, next I will drive into the road, you are a traffic information extractor." (2) "You are a real-time traffic information provider. I will ask you questions and you extract the traffic information I want from the text" (3) "I want to go to Road 53. It takes about ten minutes to drive there." Therefore, the text-to-demand agent is able to feedback: (1) Destination: You want to go to Road 53. (2) Travel time estimate: It takes about ten minutes to drive there.
Subsequently, our traffic prediction agents will use the latest 1-hour data to predict future 10-minute data. After the next 10 minutes of traffic on Road 53 are predicted, our system will utilize this data to make some suggestions using a suggestion agent, such as planning an optimal path or a traffic rush hour alert. 

\vspace{3pt} \noindent \textbf{Visualization.} Different functions of the suggestion agent need different results of traffic analysis.
For example, the optimal road planning function needs the input of short-term future traffic and then uses Dijkstra to choose the optimal road; the function of traffic alert needs short-term or long-term future traffic as input; and the road construction function needs traffic estimation of unseen roads. Additionally, in the suggestion agent, we have employed the dispatching algorithm~\cite{zhu2016Kybernetika} to realize the function of dispatching traffic police. This algorithm needs the input of short-term or long-term future traffic. We proposed a visualization scheme to present these functions, as depicted in Fig.~\ref{fig:vis_agent}.  

\begin{table*}[htb]
\caption{Datasets Size and Description}
\renewcommand\arraystretch{1}
\centering
\small
\label{Dataset_size}
\scalebox{0.95}[0.95]{$
\begin{tabular}{c|c|c|c|c|c}
\hline
\textbf{Dataset} & \textbf{Aliyun} & \textbf{PeMs-bay} & \textbf{PeMs04} & \textbf{PeMs08} & \textbf{TAXIBJ} \\ \hline
\textbf{Type} & Travel time & Speed & Flow & Flow & Flow \\ \hline
\textbf{\#-Processed Node} & 132 & 325 & 307 & 170 & 1024 \\ \hline
\textbf{\#-Edge} & 420 & 2694 & 680 & 548 & 4114 \\ \hline
\textbf{Time step} & 51240 & 52116 & 16992 & 17865 & 3600 \\ \hline
\textbf{Slice} & 2 min & 5 min & 5 min & 5 min & 1 hour \\ \hline
\end{tabular}$}
\end{table*}

\section{Experiments}
\label{Sec-Experiments}


\subsection{Experimental Settings}
To evaluate the performance of the framework, we conduct experiments in four real datasets. Three agent modules are implemented in \texttt{Python 3.7} based on \texttt{PyTorch}.

\vspace{3pt} \noindent \textbf{Datasets:} To comprehensively evaluate our system, we use the following five real-world datasets in our experiments: \textbf{Aliyun}\footnote{https://tianchi.aliyun.com/competition/entrance/231598/information?from=oldUrl}, \textbf{PeMs-bay}~\cite{LIiclr2018}, \textbf{PeMs04}~\cite{Guoaaai2019}, \textbf{PeMs08}~\cite{Guoaaai2019}, \textbf{TAXIBJ}~\cite{pan2020spatio}. Table~\ref{Dataset_size} summarises the statistical information of datasets and detailed information on datasets is as follows:
\begin{itemize}
 \item \textbf{Aliyun} contains 132 roads, 4 attributes, and about 7.67 million data records from Guizhou in China, ranging from 1st March to 30th June 2017. Each road contains 720 data records per day and the time slice is 2 minutes. We selected 120 roads interactive roads to establish the Aliyun dataset.
 
\item \textbf{PeMS-bay} is a public traffic speed dataset collected from California Transportation Agencies (CalTrans) Performance Measurement System (PeMs). PeMs-bay contains 6 months of data recorded by 325 traffic sensors ranging from January 1st, 2017 to June 30th, 2017 in the Bay Area.

\item \textbf{PeMs04} is a public traffic flow dataset collected from CalTrans PeMS. Specifically, PEMS04 contains data from 307 sensors in District04 over a period of 2 months from Jan 1st 2018 to Feb 28th 2018. 
 
\item \textbf{PeMs08} is a public traffic flow dataset collected from CalTrans PeMS. Specifically, PEMS08 contains data from 170 sensors in District08 over a period of 2 months from July 1st 2018 to Aug 31th 2018.

\item \textbf{TAXIBJ} is a public taxi flow dataset collected from Beijing city, which contains a large number of taxicab trajectories from Feb. 1st 2015 to Jun. 2nd 2015, and 690242 roads with 8 attributes, including length, width, the number of lanes, etc. This dataset is utilized in scalability experiments.
\end{itemize}

\vspace{3pt} \noindent \textbf{Evaluation Metrics}. We utilize the following three metrics to evaluate the performance: $RMSE=\sqrt{\frac{1}{M}\sum_{i=1}^M {(y_i-\hat{y}_i)^2}}$, $MAE=\frac{1}{M}\sum_{i=1}^M{|y_i-\hat{y}_i|}$, $MAPE=\frac{100\%}{M}\sum_{i=1}^M{|\frac{y_i-\hat{y}_i}{y_i}|}$, where $y_i$ and $\hat{y}_i$ represent the real value and predicted value of the test sample $i$, respectively. Let $M$ be the number of test samples.

\vspace{3pt} \noindent \textbf{Baselines}. We compared the performance of \textsf{TrafficGPT} with the following rivals: 


\begin{itemize}
   \item \textbf{STGCN~\cite{YuIJCAI2018}}: The Spatial-Temporal Graph Convolutional Networks also model spatial and temporal dependencies with several spatial-temporal graph convolutional blocks. We set the channel number of the first, and second spatial-temporal graph convolution blocks of the STGCN model as 8, and 16, respectively.
   \item \textbf{STSGCN~\cite{songaaai2020}}: STSGCN is proposed to effectively capture
    the localized spatial-temporal correlations and consider the
    heterogeneity in spatial-temporal data.
    \item \textbf{GMAN~\cite{zhengaaai2020}}:  GMAN is an attention-based model which
    stacks spatial, temporal and transform attentions.
    \item \textbf{STTN~\cite{xuarxiv2020}}: STTN is an attention-based deep network that leverages dynamical spatial-temporal dependencies.
\end{itemize}

We rigorously evaluated our proposed \textsf{TrafficGPT} system through three sets of experiments. Initially, we compared \textsf{TrafficGPT} against multiple baselines in short-term prediction tasks on four widely-used traffic datasets, to evaluate the important accuracy performance of our system. Next, we conducted long-term prediction experiments on four datasets, showcasing the robust autoregressive ability and scalability of our system. Finally, we assessed the performance of \textsf{TrafficGPT} on the unseen road predicting tasks, using TAXIBJ and Aliyun datasets, analyzing its performance with larger datasets and unseen road segments.

\begin{figure*}[htb]
	\begin{center}
            \begin{tabular}{cccc}
			\includegraphics[width=0.45\columnwidth]{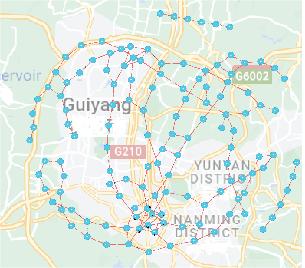}&\includegraphics[width=0.45\columnwidth]{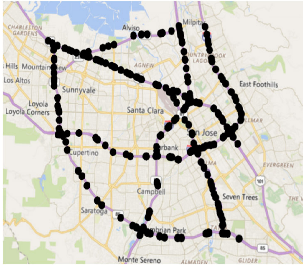}&
            \includegraphics[width=0.45\columnwidth]{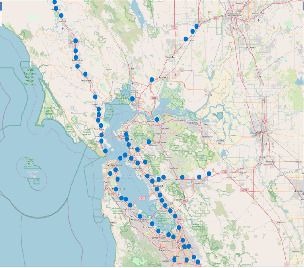}&
			\includegraphics[width=0.45\columnwidth]{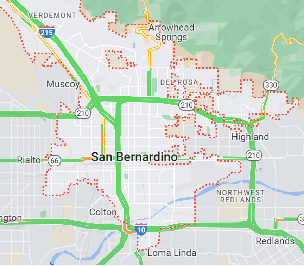}
			
			  \\
                {\scriptsize (a) Road network of Aliyun}& {\scriptsize (b) Road network of PeMs-bay}&
			{\scriptsize (c) Road network of PeMs04}& {\scriptsize (d) Road network of PeMs08}  \\
		\end{tabular}
		
            \begin{tabular}{cccc}
			\includegraphics[width=0.45\columnwidth]{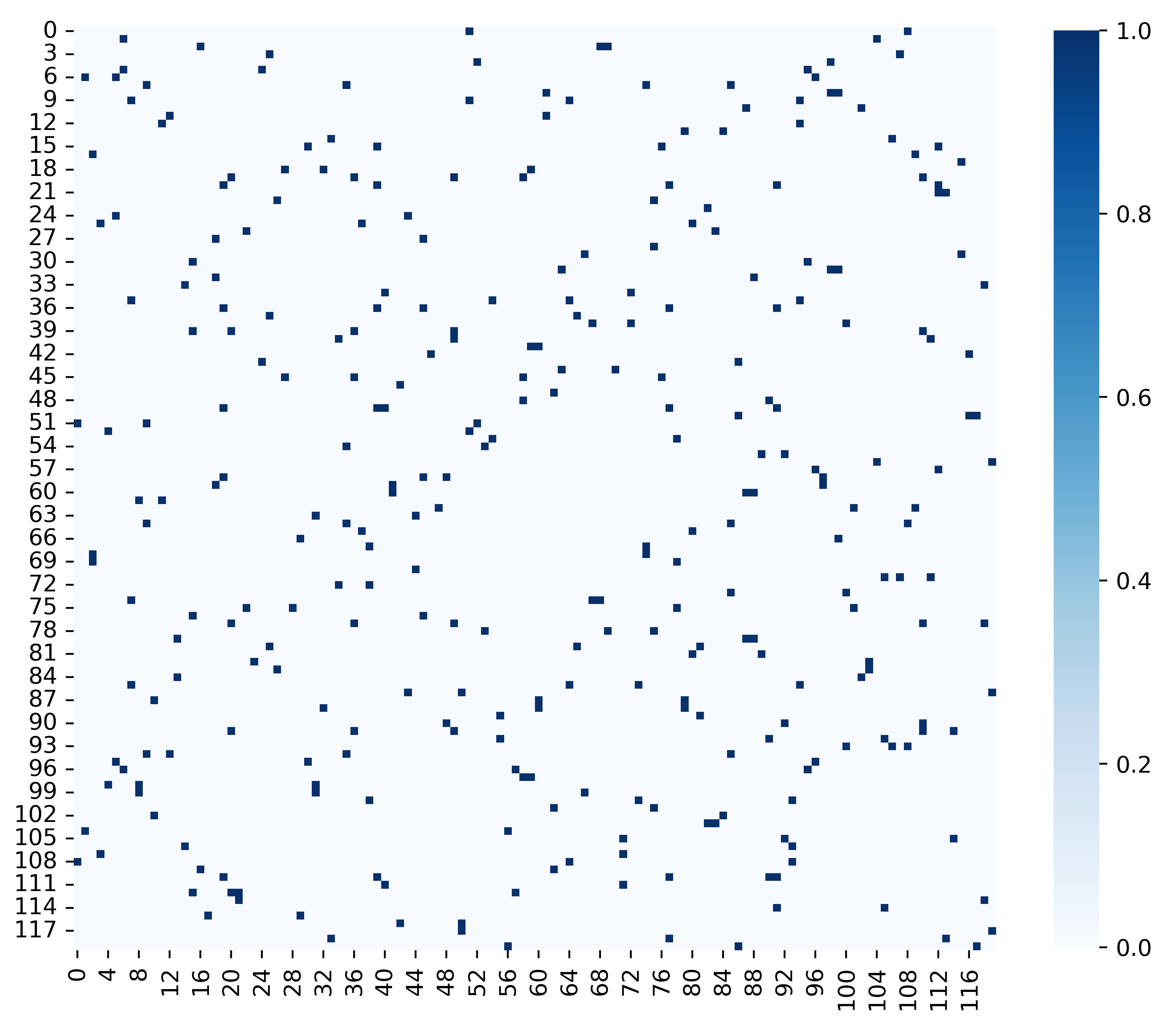}&\includegraphics[width=0.45\columnwidth]{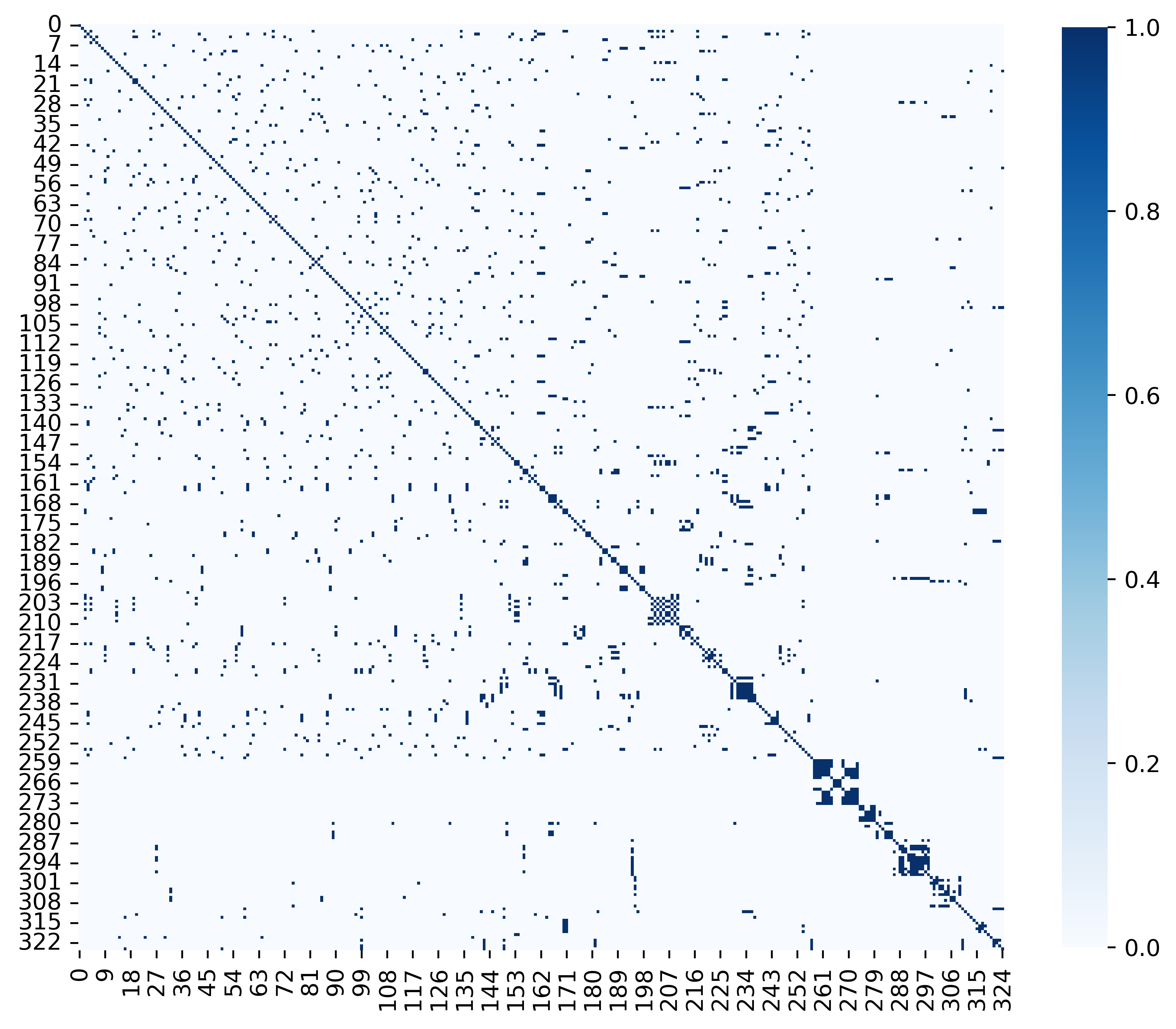}&
            \includegraphics[width=0.45\columnwidth]{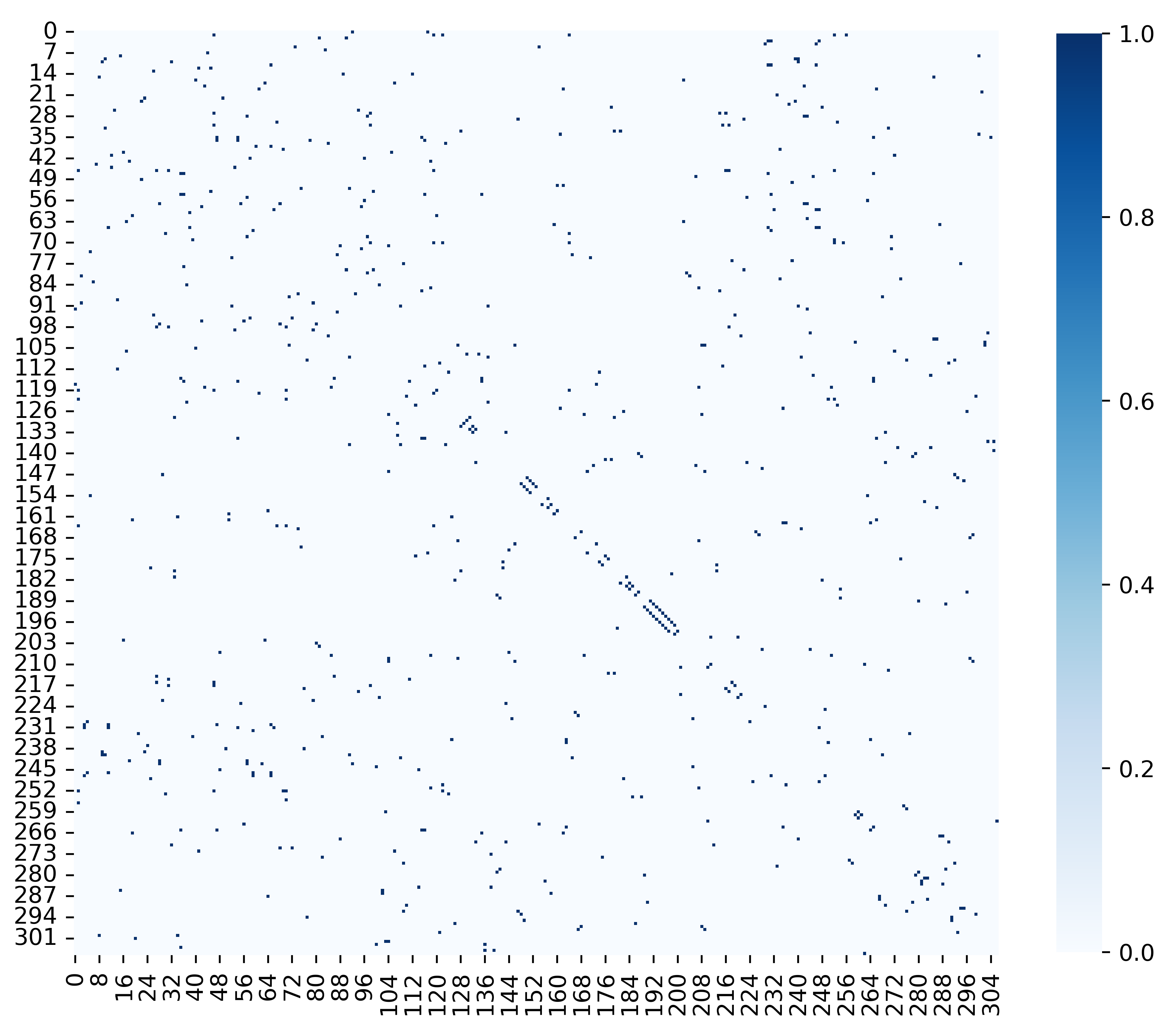}&
			\includegraphics[width=0.45\columnwidth]{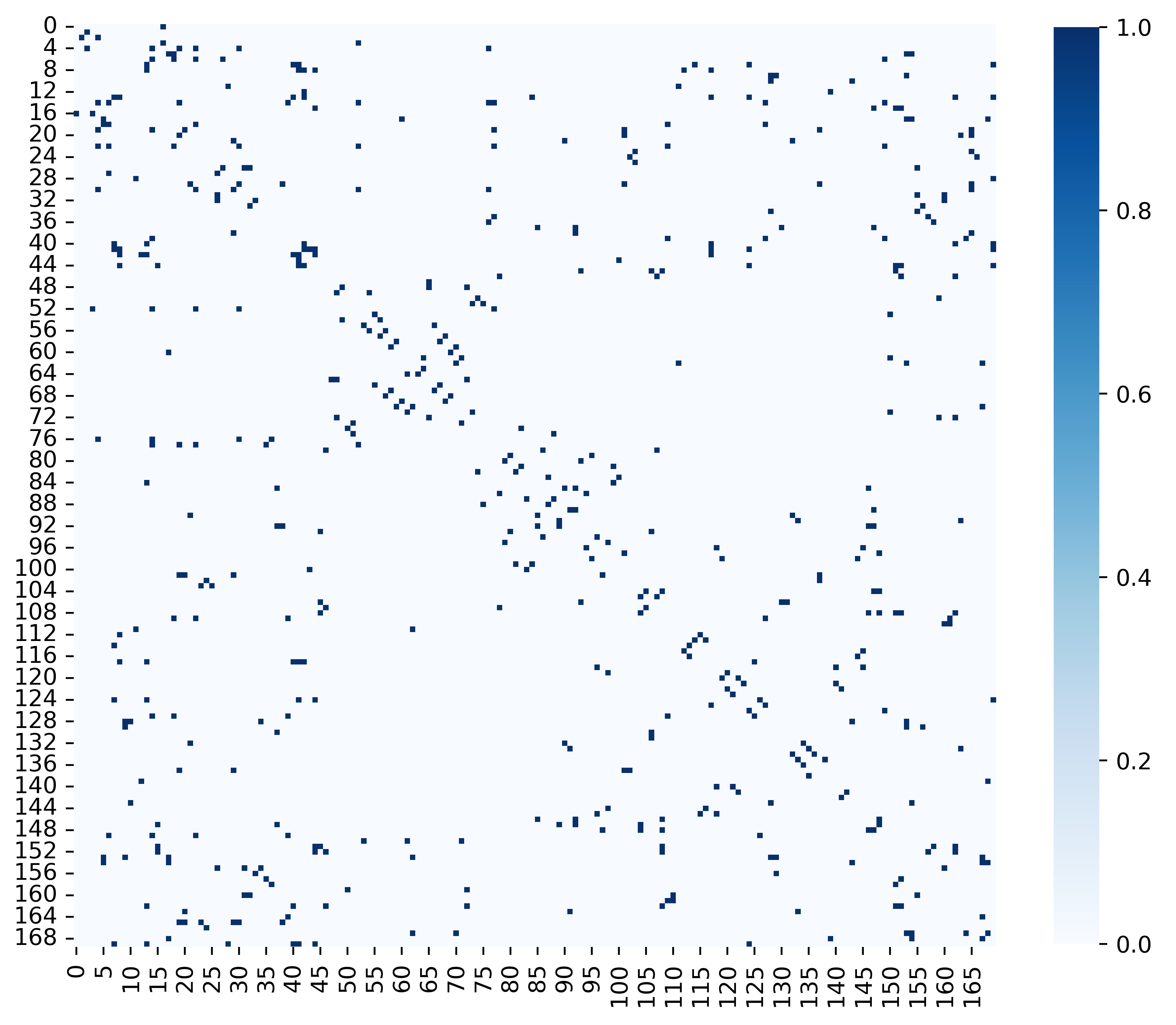}
			
			  \\
                {\scriptsize (e) Connectivity graph of Aliyun}& {\scriptsize (f) Connectivity graph of PeMs-bay}&
			{\scriptsize (g) Connectivity graph of PeMs04}& {\scriptsize (h) Connectivity graph of PeMs08}  \\
		\end{tabular}

            \begin{tabular}{cccc}
			\includegraphics[width=0.45\columnwidth]{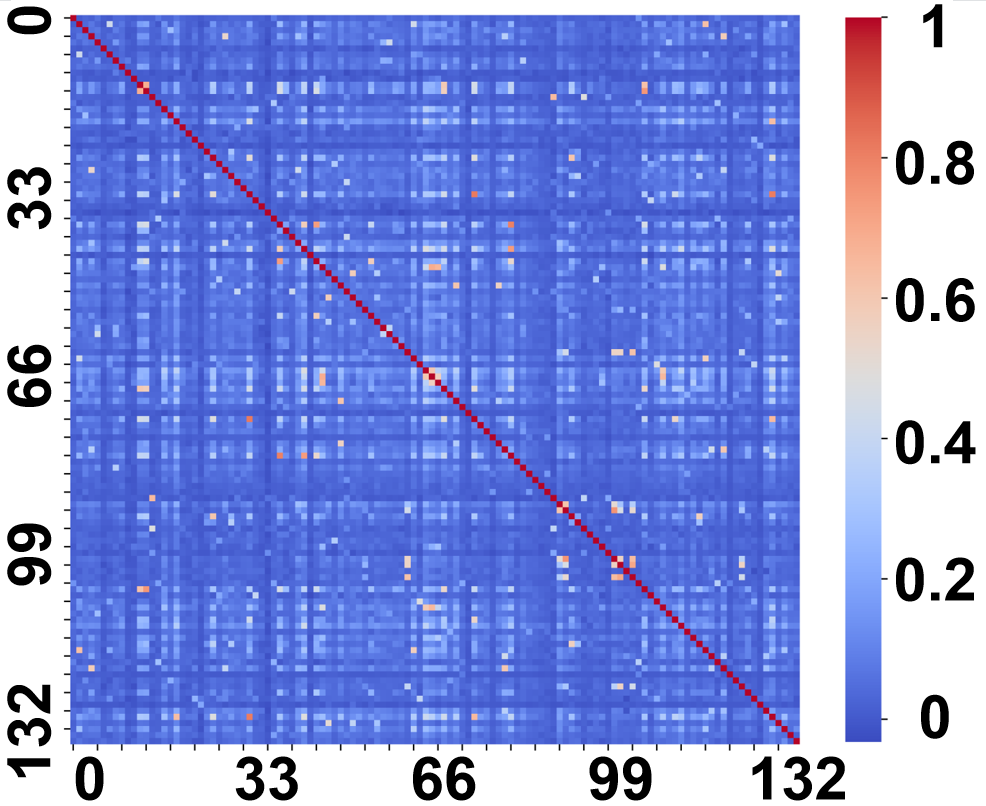}&\includegraphics[width=0.45\columnwidth]{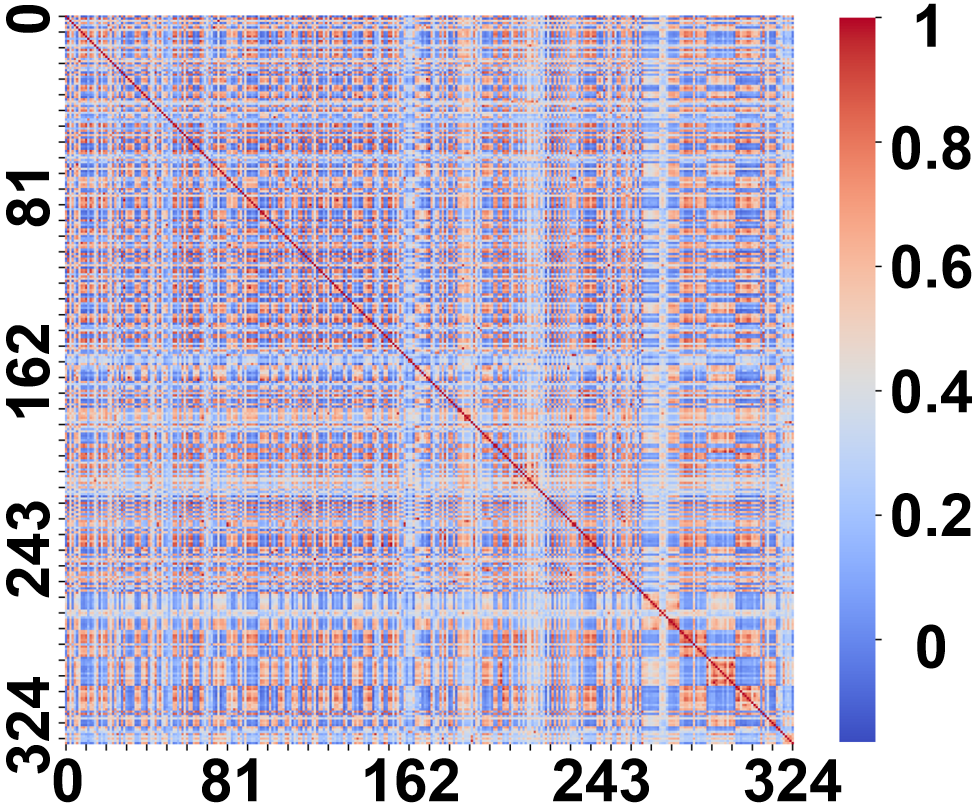}&
            \includegraphics[width=0.45\columnwidth]{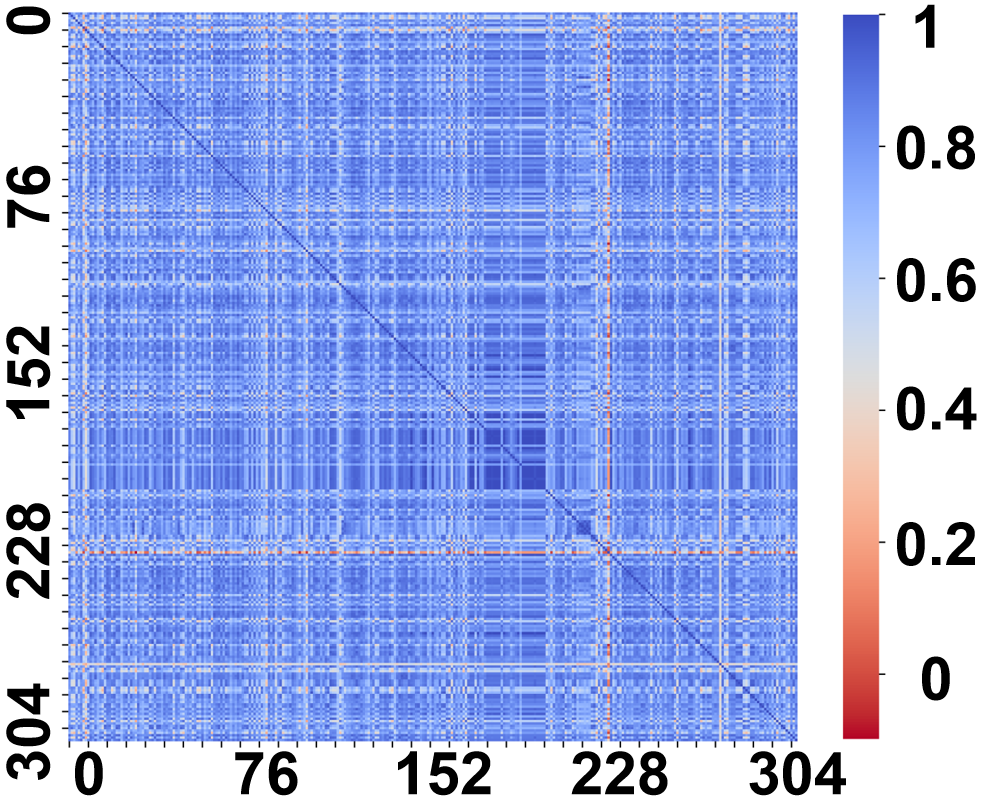}&
			\includegraphics[width=0.45\columnwidth]{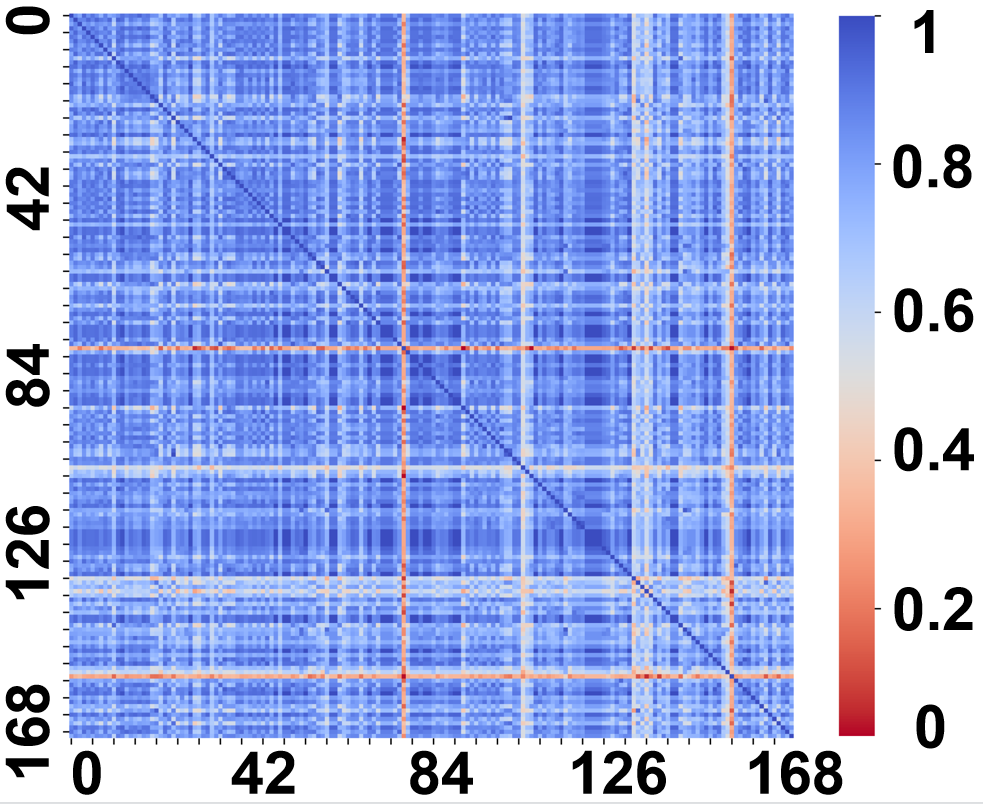}
			
			  \\
                {\scriptsize (i) Correlation graph of Aliyun}& {\scriptsize (j) Correlation graph of PeMs-bay}&
			{\scriptsize (k) Correlation graph of PeMs04}& {\scriptsize (l) Correlation graph of PeMs08}  \\
		\end{tabular}

            \begin{tabular}{cccc}
			\includegraphics[width=0.45\columnwidth]{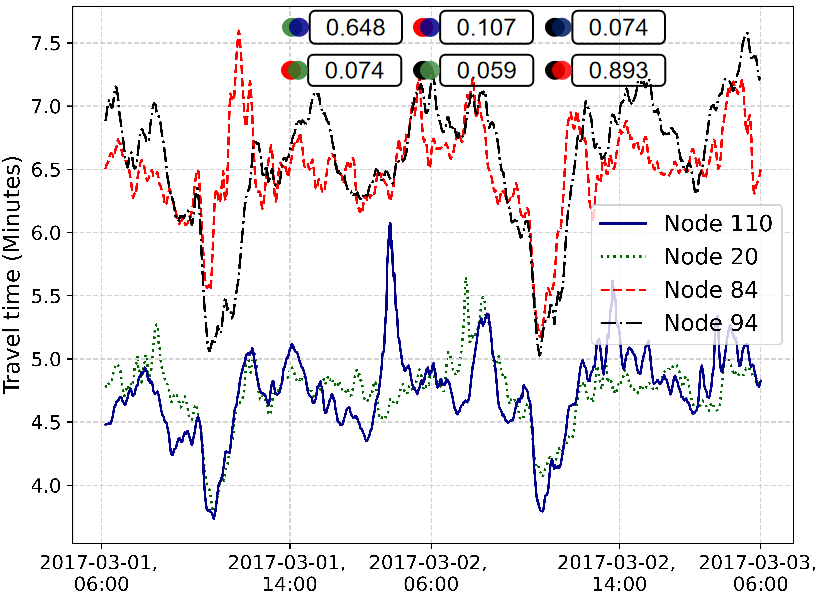}&\includegraphics[width=0.45\columnwidth]{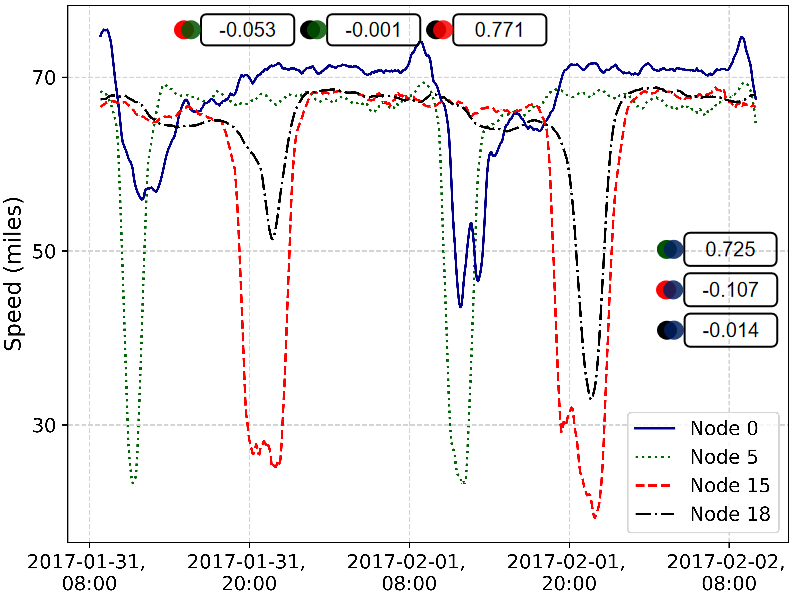}& \includegraphics[width=0.45\columnwidth]{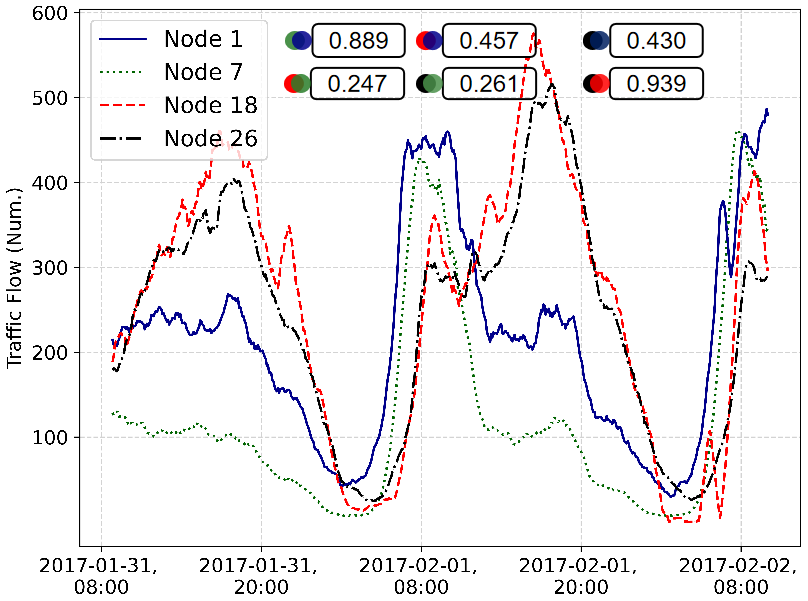}&\includegraphics[width=0.45\columnwidth]{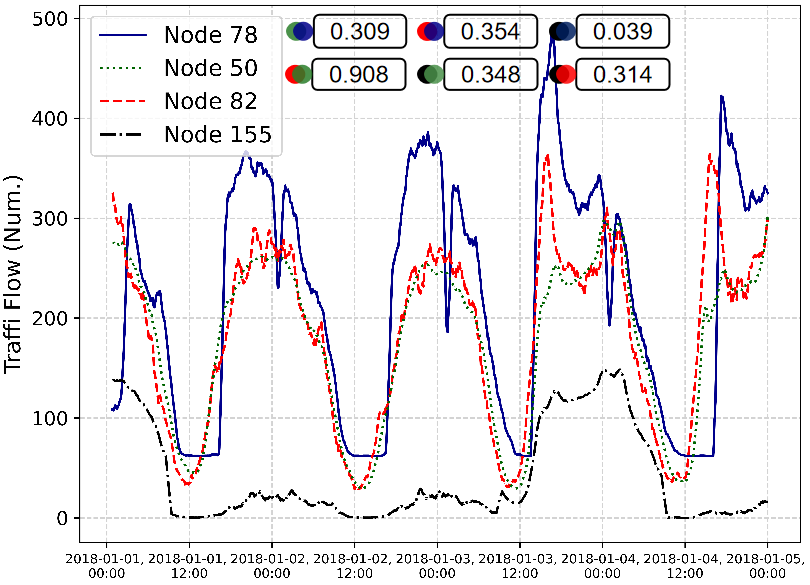}\\
			{\scriptsize (o) Traffic flow on Aliyun.} & {\scriptsize (p) Traffic flow on PeMs-bay.}&{\scriptsize (q) Traffic flow on PeMs04.} & {\scriptsize (r) Traffic flow on PeMs08.}
		\end{tabular}
	\end{center}
	\caption{Road network, connectivity graph, and correlation graph of four datasets.}
	\label{fig:corr_all_2}
\end{figure*}

\subsection{Performance of Short-term Prediction }
We use four datasets (Aliyun, PeMs-bay, PeMs04, and PeMs08) of complex road networks to evaluate the performance of short-term prediction.

\textbf{Hyperparameter.} 
We use 12 historical time steps (1 hour) on PeMs-bay, PeMs04, and PeMs08 datasets to predict the traffic conditions of the next $Q = 12$ steps (1 hour). Since the different recording rates between Aliyun and PeMs datasets, we use 30 historical time steps (1 hour) on Aliyun dataset. We train our model using Adam optimizer with an initial learning rate of 0.001 and using the $L_1$-norm loss function. In the spatial and temporal transformers, we set 4, 16, 16, and 16 heads, respectively. The number of spatial-temporal blocks is 2 and the embed size is 64. Besides, the $X_w$ and $X_d$ lengths are set to 3 by default.

\subsubsection{Correlation Analysis}
To illustrate the effectiveness of correlation representation inside the \textsf{TrafficGPT}, we conducted some analysis on four datasets. As shown in Fig.~\ref{fig:corr_all_2}(a-d) and (e)-(h), Aliyun and PeMs-bay networks are ring roads, whereas PeMs04 and PeMs08 networks are majorly straight structures. It is foreseeable that each node inside the network of PeMs04 or PeMs08 would have a stronger correlation with other nodes inside the network as illustrated in Fig.~\ref{fig:corr_all_2}(i)-(l). The upstream road would directly influence the downstream road in the straight structural road network due to the unique connectivity. In contrast, ring road networks such as Aliyun and PeMs-bay would present a more complex correlation between nodes. To further ensure, we select some period time of nodes and then visualize the traffic conditions to verify it. To show TrafficGPT's ability to discriminate between similar traffic data, we smoothed the data to demonstrate it. As illustrated in Fig.~\ref{fig:corr_all_2}(o)-(-r), the correlation matrix is higher, and the traffic conditions are more similar on the four datasets.

\subsection{Performance of Long-term Prediction}
\label{sec:long-term:exp}
In this section, we have evaluated the performance of our framework in the long-term prediction task on four datasets; meanwhile, to ensure our work is scalable to a larger road network, we have conducted a scalability evaluation on the TAXIBJ dataset.
\begin{figure*}[tb]
	\begin{center}
		\begin{tabular}{cccc}
			
            \includegraphics[width=0.45\columnwidth]{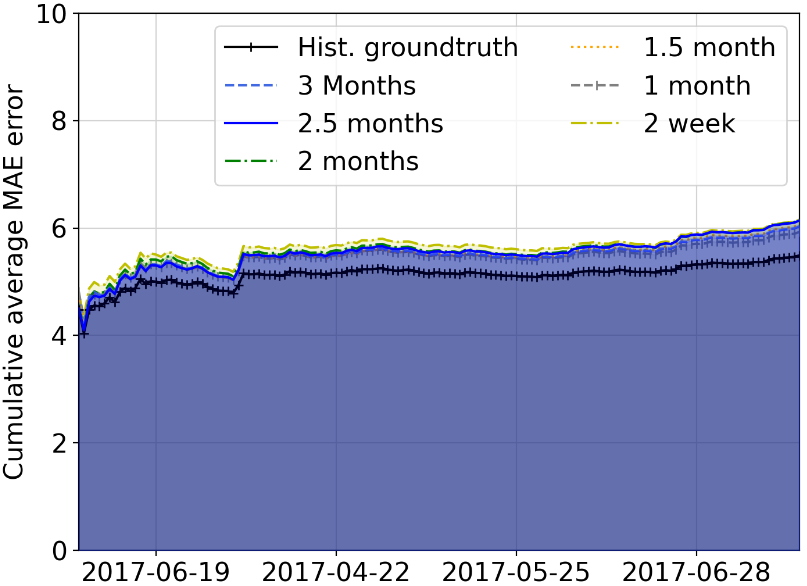}&
			\includegraphics[width=0.45\columnwidth]{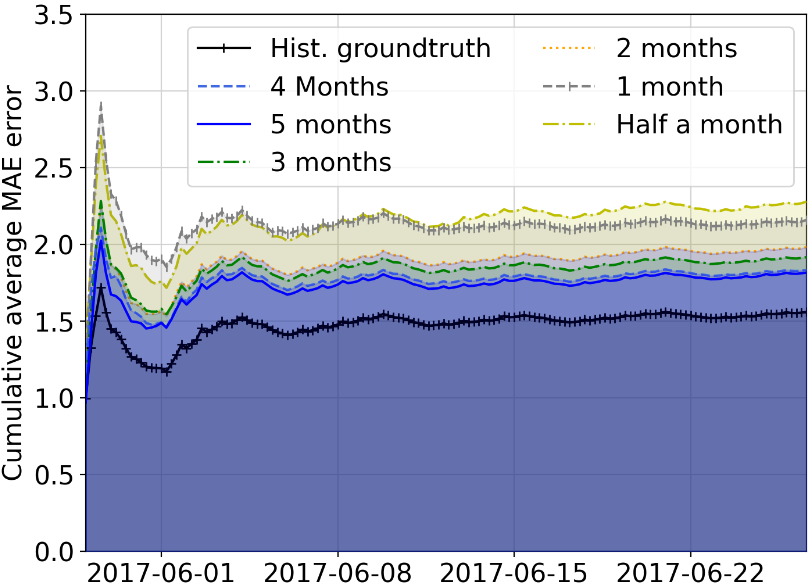}&
			\includegraphics[width=0.45\columnwidth]{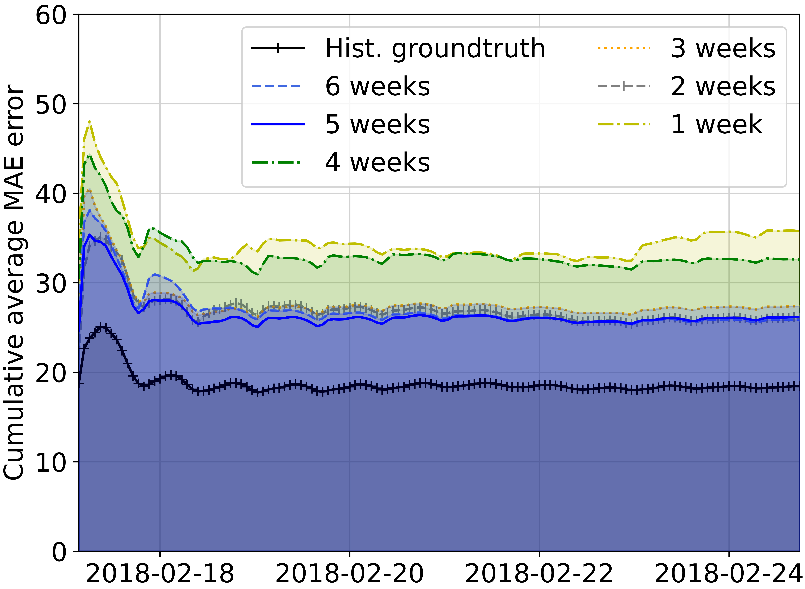}&
			\includegraphics[width=0.45\columnwidth]{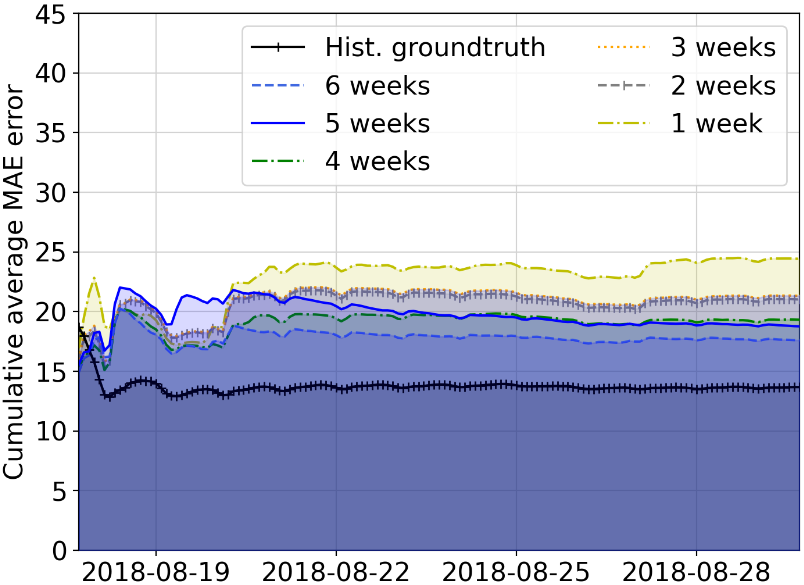}\\
			{\scriptsize (a) Aliyun.}& {\scriptsize (b) PeMs-bay.} & {\scriptsize (c) PeMs04.} & {\scriptsize (d) PeMs08.} \\
		\end{tabular}
	\end{center}
	\caption{Performance of long-term prediction.}
	\label{fig:longterm results}
\end{figure*}

\subsubsection{Long-term Prediction Performance}
We used the autoregressive methods mentioned in Section~\ref{sec:long-term method} to fine-tune our model, then used the beginning 1-hour data of four datasets as input, to conduct the long-term predictions. To evaluate the influence of the well-trained model (mentioned in Section~\ref{sec:long-term method}), we set different lengths of training data to well-train the short-term model at first, followed by fine-tuning. The results are shown in Fig.~\ref{fig:longterm results}. The results show that compared with short-term prediction results (marked as Hist. groundtruth), the average cumulative MAE increases more on the datasets of PeMs04 and PeMs08; while, on the datasets of Aliyun and PeMs-bay, \textsf{TrafficGPT} achieved superior performance with different lengths of training data. The results indicate that the long-term prediction of our system performs better in the more complex road networks. However, in the more simple networks, accidental events influence the entire network deeply, which directly reduces the predicting accuracy. 

Moreover, the MAEs of long-term prediction on four datasets are stable during the prediction, which indicates no error explosion. This implies that even though the accuracy suffers reduction, our system can predict long-term traffic conditions with acceptable precision, which is highly informative.

\subsubsection{Scalability Analysis}
\label{sec:scalability}
Furthermore, to test the stability and robustness of our framework, we conducted experiments using top-performing methods on a larger dataset called TAXIBJ containing 690,242 roads, which split Beijing city into a 32×32 grid, whose statistics are in Table~\ref{taxi flow}. We partitioned and scaled the road network by nodes and trained TrafficGPT, GMAN, STSGCN, STTN, and GWN. We evaluated our framework's scalability by selecting 20\%, 40\%, 60\%, 80\%, and 100\% of the nodes in TAXIBJ, using either partitioned or complete data. Additionally, to improve our framework's ability to adapt to new road scenarios quickly, we implemented a road network mapping evaluation approach that includes a fine-tuning stage. During the evaluation, we trained models using 12 hours of historical data (12 time slices) and predicted the future 3 hours of data (3 time slices).

\begin{table}[htb]
 \caption{Extensive details of the TAXIBJ datasets.}
\renewcommand\arraystretch{0.9}
\centering
\small
\label{taxi flow}
\scalebox{0.9}[0.9]{$
\begin{tabular}{l|l|c}
\hline
Datasets                       & Property           & \multicolumn{1}{l}{Value} \\ \hline
\multirow{2}{*}{POIs}          & \# POIs            & 982,829                   \\
                               & \# POI categories  & 668                       \\ \hline
\multirow{2}{*}{Road networks} & \# roads           & 690,242                   \\
                               & \# road attributes & 8                         \\ \hline
\multirow{4}{*}{*Geo-graph}    & \# nodes           & 1024                      \\
                               & \# edges           & 4114                      \\
                               & \# node features   & 989                       \\
                               & \# edge features   & 32                        \\ \hline
\end{tabular}
$}
\end{table}

We obtained taxi flow data from the TDrive dataset, containing taxicab trajectories from Feb. 1st, 2015 to Jun. 2nd 2015. Hourly inflows and outflows were extracted for each grid by tallying the number of taxis entering or exiting the grid.

Geo-graph attributes were derived from Points of Interest (POIs) and road networks (RNs) in Beijing, consisting of 982,829 POIs belonging to 668 categories and 690,242 roads with 8 attributes, including length, width, and number of lanes. Node attributes for each grid included POI and RN features such as category counts, number of roads, and lanes. Edge attributes described road features connecting pairs of grids, including the number of roads and lanes.

\subsubsection{Scalability Evaluation.} 
\begin{figure}[tb]
	\begin{center}
		\begin{tabular}{cc}		
            \includegraphics[width=0.45\columnwidth]{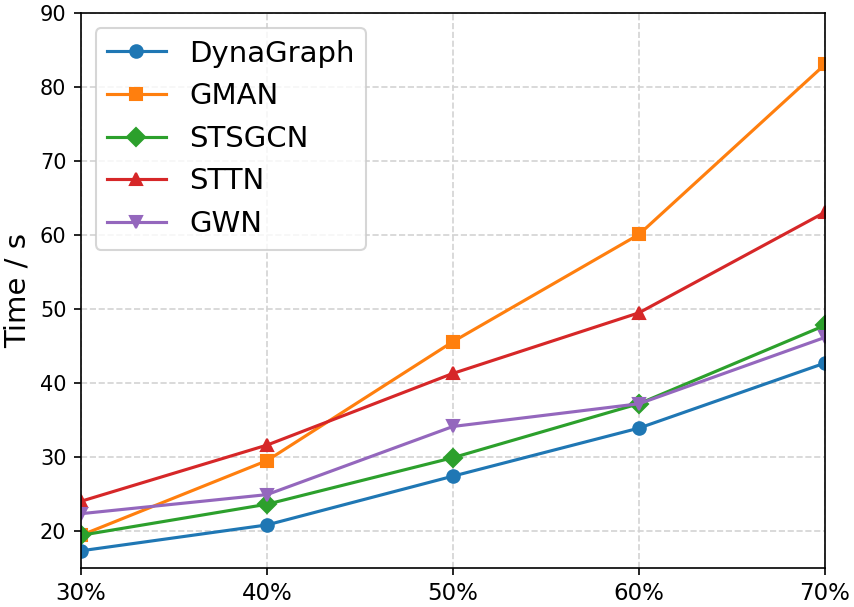}&
			\includegraphics[width=0.43\columnwidth]{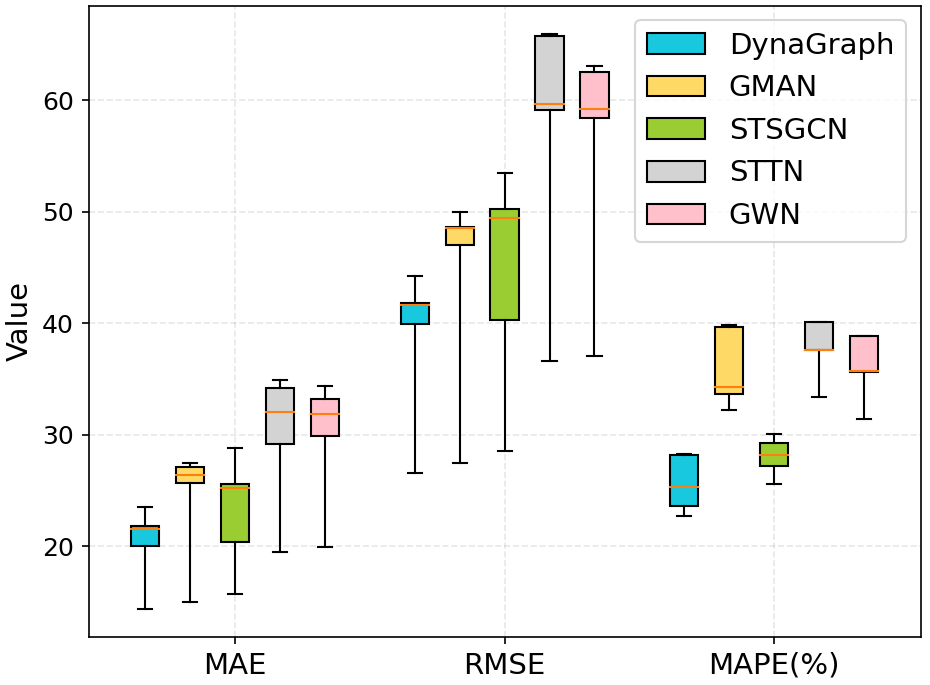}\\
			{\scriptsize (a) One epoch training time range on scales.}& {\scriptsize (b) Accuracy variations range on scales.}  \\
		\end{tabular}
	\end{center}
	\caption{Scalability analysis.}
	\label{fig:scale_all}
\end{figure}

To validate the effectiveness of our framework in achieving accurate predictions for any sizable road networks, we conducted a scalability experiment to demonstrate the prediction precision of our framework. The results are summarized in Table~\ref{tab:scalability_exp2}, which indicates that our framework outperforms other methods across different scaled sizes of TaxiBJ. Additionally, we analyzed the training time of one epoch, as shown in Fig.~\ref{fig:scale_all} (a). It can be observed that our framework exhibits approximate linear growth as the scaled size increases. In comparison with GMAN, STSGCN, STTN, and GWN, our framework performs better in terms of time efficiency, demonstrating its scalability. Besides, TrafficGPT exhibits superior performance on various scaled road networks, as evident from Table~\ref{tab:scalability_exp2} and Fig.~\ref{fig:scale_all} (b).

\begin{table*}[htb]
        \begin{minipage}{1\textwidth}
        \scriptsize
    \centering
    \caption{Preliminary experiment of DMFT model on TAXIBJ}
    \scalebox{1}[1]{
    \begin{tabular}{|c|*{15}{c|}}
\hline
\multirow{2}{*}{Scale} & \multicolumn{3}{c|}{TrafficGPT} & \multicolumn{3}{c|}{GMAN} & \multicolumn{3}{c|}{STSGCN} & \multicolumn{3}{c|}{STTN} & \multicolumn{3}{c|}{GWN} \\
\cline{2-16}
& MAE & RMSE & MAPE & MAE & RMSE & MAPE & MAE & RMSE & MAPE & MAE & RMSE & MAPE & MAE & RMSE & MAPE \\
\hline
100\% & \textbf{20.02} & 41.86 & 28.26\% & 25.66 & 49.94 & 39.88\% & 20.34 & \textbf{40.30} & \textbf{28.16\%} & 29.14 & 59.09 & 40.12\% & 29.89 & 59.19 & 38.84\% \\
\hline
80\% & \textbf{21.82} & \textbf{41.60} &\textbf{ 25.34\%} & 27.15 & 48.58 & 39.65\% & 25.23 & 49.46 & 27.23\% & 34.18 & 65.93 & 37.61\% & 33.17 & 62.50 & 35.64\% \\
\hline
60\% & \textbf{23.49} & \textbf{44.25} & \textbf{23.64\%} & 27.46 & 48.62 & 33.63\% & 28.77 & 53.50 & 29.26\% & 34.94 & 65.77 & 33.42\% & 34.35 & 63.06 & 35.73\% \\
\hline
40\% & \textbf{21.55} & \textbf{39.98} & \textbf{22.67\%} & 26.40 & 46.99 & 32.19\% & 25.56 & 50.29 & 25.61\% & 32.04 & 59.71 & 32.24\% & 31.86 & 58.38 & 31.42\% \\
\hline
20\% & \textbf{14.40} & \textbf{26.59} & \textbf{28.15\%} & 15.04 & 27.51 & 34.29\% & 15.75 & 28.51 & 30.06\% & 19.46 & 36.63 & 36.93\% & 19.89 & 37.08 & 37.94\% \\
\hline
\end{tabular}}
\label{tab:scalability_exp2}
    \end{minipage}%
\end{table*}

\begin{table*}[htb]
    \begin{minipage}{1\textwidth}
        \centering
\scriptsize
\caption{Road mapping evaluation results}
\scalebox{1}[1]{$
\begin{tabular}{|l|l ll|lll|lll|lll|}
\hline
Training data & \multicolumn{3}{c|}{10\% num. of data} & \multicolumn{3}{c|}{20\% num. of data} & \multicolumn{3}{c|}{50\% num. of data} & \multicolumn{3}{c|}{100\% num. of data} \\ \hline
{Metrics} & \multicolumn{1}{l|}{MAE} & \multicolumn{1}{l|}{RMSE} & MAPE & \multicolumn{1}{l|}{MAE} & \multicolumn{1}{l|}{RMSE} & MAPE & \multicolumn{1}{l|}{MAE} & \multicolumn{1}{l|}{RMSE} & MAPE & \multicolumn{1}{l|}{MAE} & \multicolumn{1}{l|}{RMSE} & MAPE \\ \hline
100\% & \multicolumn{1}{l|}{21.19} & \multicolumn{1}{l|}{43.23} & 36.50\% & \multicolumn{1}{l|}{18.32} & \multicolumn{1}{l|}{37.38} & 32.05\% & \multicolumn{1}{l|}{16.73} & \multicolumn{1}{l|}{34.26} & 27.48\% & \multicolumn{1}{l|}{16.46} & \multicolumn{1}{l|}{33.69} & 25.10\% \\ \hline
80\% & \multicolumn{1}{l|}{23.72} & \multicolumn{1}{l|}{45.29} & 32.63\% & \multicolumn{1}{l|}{20.86} & \multicolumn{1}{l|}{40.19} & 28.53\% & \multicolumn{1}{l|}{19.42} & \multicolumn{1}{l|}{37.66} & 25.00\% & \multicolumn{1}{l|}{19.04} & \multicolumn{1}{l|}{37.34} & 22.60\% \\ \hline
60\% & \multicolumn{1}{l|}{24.99} & \multicolumn{1}{l|}{46.71} & 30.00\% & \multicolumn{1}{l|}{22.20} & \multicolumn{1}{l|}{42.10} & 27.10\% & \multicolumn{1}{l|}{20.86} & \multicolumn{1}{l|}{40.03} & 24.11\% & \multicolumn{1}{l|}{20.67} & \multicolumn{1}{l|}{39.54} & 21.87\% \\ \hline
40\% (disjoint) & \multicolumn{1}{l|}{12.98} & \multicolumn{1}{l|}{31.04} & 39.01\% & \multicolumn{1}{l|}{11.29} & \multicolumn{1}{l|}{26.44} & 35.82\% & \multicolumn{1}{l|}{10.80} & \multicolumn{1}{l|}{24.74} & 32.11\% & \multicolumn{1}{l|}{10.72} & \multicolumn{1}{l|}{24.50} & 32.86\% \\ \hline
\end{tabular}$}
\label{tab:road_map_exp2}
    \end{minipage}
\end{table*}

\subsection{Performance of Unseen Road Estimation}
In this section, we use two types of tasks to evaluate the ability, of the third agent, to predict unseen roads. The first experiment is to test the performance in network expansion; the second experiment is to test the performance of predicting newly added roads.

\subsubsection{Road Network Mapping Evaluation.}
This is a preliminary experiment that we preliminary test the feasibility of estimating unseen roads traffic, which are under the same road network. To validate the capability of our framework in handling road network expansion tasks, we incorporated a two-stage for the road mapping evaluation on TAXIBJ dataset. During the first preliminary stage, we utilized 40\% of the nodes data to pre-train our framework and test on another percentage of all nodes. To address the mismatch in dimensional input and output data across different scaled sizes of road networks, we introduced three linear layers to downscale the input features $X_c$, $X_d$, and $X_w$ in terms of node dimensions (e.g., downsizing from nodes = 810 to nodes = 415, matching the dimensions of the 40\% pre-trained framework input). Subsequently, a linear layer was appended at the end to upscale the node dimensions (e.g., upsizing from nodes = 415 to nodes = 810).

In the second fine-tuning stage, we initialized framework using the parameters from pre-trained framework in first stage. Then, we conducted the second fine-tuning process to fit our framework, subsequently predict these nodes' traffic. 

\begin{figure}[tb]
	\begin{center}
		\begin{tabular}{cc}
			
            \includegraphics[width=0.45\columnwidth]{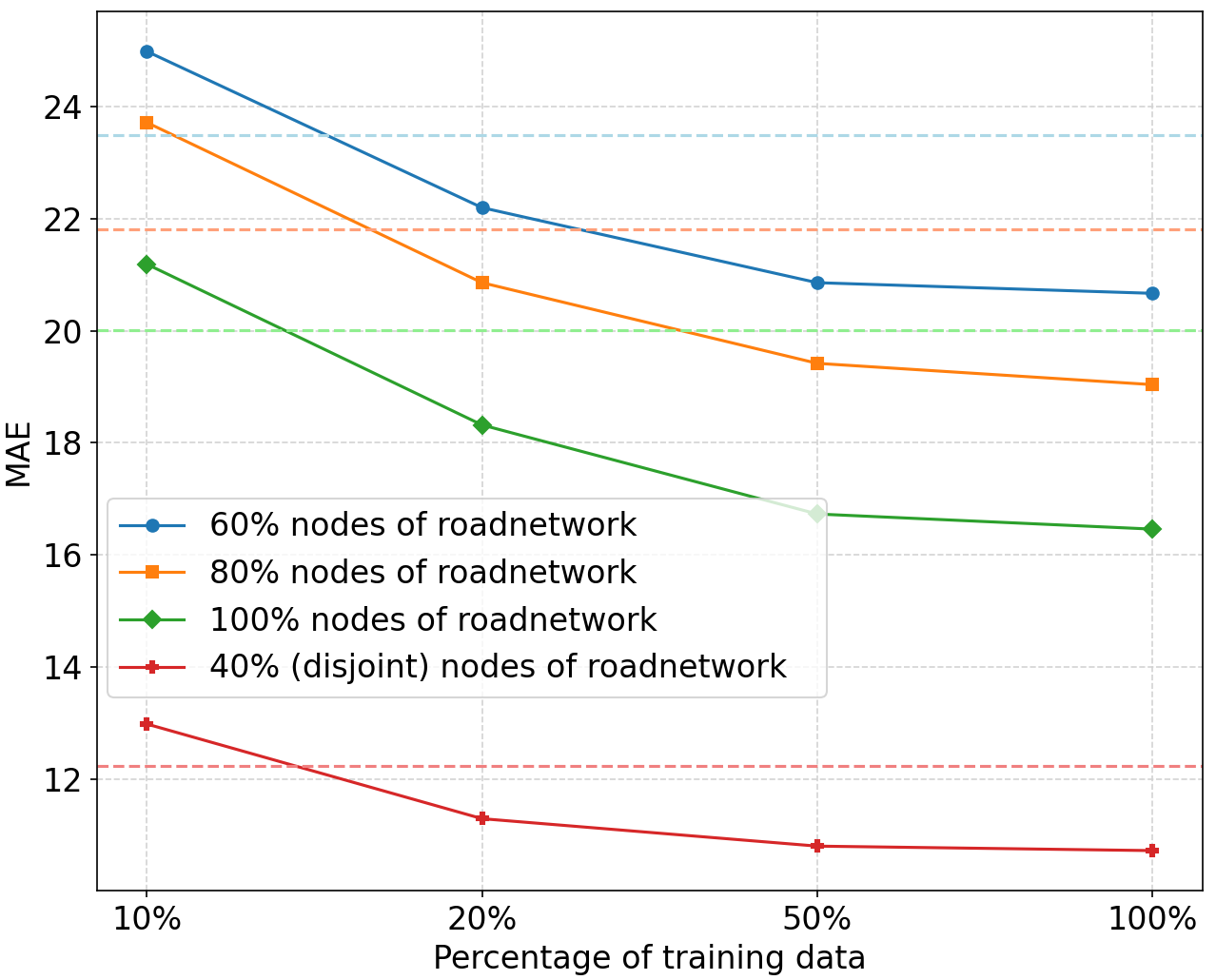}&
			\includegraphics[width=0.45\columnwidth]{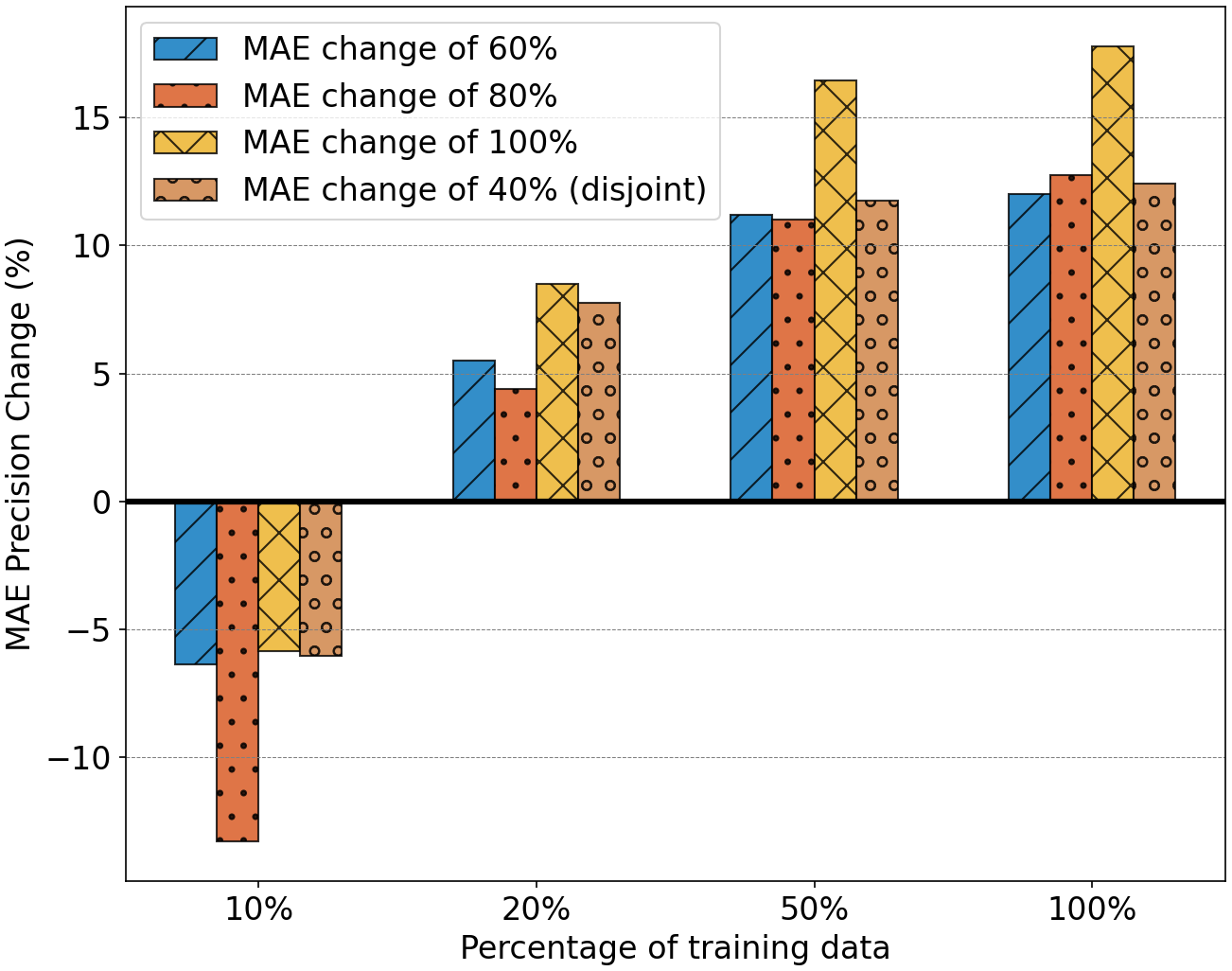}\\
			{\scriptsize (a) Change with training data size.}& {\scriptsize (b) fine-tuned vs. direct-fitted.}  \\
		\end{tabular}
	\end{center}
	\caption{Mapping evaluation analysis with MAE changes.}
	\label{fig:finetune}
\end{figure}

Specifically, we utilized 40\% of the nodes' data for first pre-training stage, and used disjoint 40\%, 60\%, 80\%, and 100\% of the nodes' data for the second fine-tuning stage. Additionally, we use a portion of these nodes' data as fine-tuning data to show the impacts of the lengths of fine-tuning data. The results are shown in Fig.~\ref{fig:finetune} and Table~\ref{tab:road_map_exp2}. The results show that using partial nodes to train our framework can predict other unseen nodes with acceptable accuracy. We observed that utilizing only 10\% of the training data for fine-tuning, especially for different or disjoint road networks, resulted in a decline in accuracy by 5\% to 12\%. However, when we increased the percentage of training data for fine-tuning to 20\%, we observed an improvement in accuracy by 4\% to 8\%. This finding was surprising as we discovered that having an adequate number of training data for new road networks can actually enhance accuracy compared to direct fitting. To further illustrate this, we plotted the changes in Mean Absolute Error (MAE) with respect to the percentage of training data in Fig.~\ref{fig:finetune} (a). The figure shows that with as little as 10\% to 20\% of training data used in the fine-tuning stage, \textsf{TrafficGPT} can achieve comparable precision performance on unseen road with directly fitted \textsf{TrafficGPT}.


This phenomenon indicates that changes in the size of the road network, whether it expands or shrinks, do not significantly impact the accuracy of our model. On the other hand, larger road networks tend to achieve higher accuracy after fine-tuning compared to direct training. This can be attributed to the fact that the parameter initialization from the pre-trained framework provides the model with a gradient convergence direction that is closer to the optimal solution, resulting in improved accuracy. This result provides significant statistical evidence for the scalability of our \textsf{TrafficGPT} framework in handling any sizable road networks.

\section{Conclusions}
\label{Sec-Con}

In this paper, we have proposed \textsf{TrafficGPT} system, a system towards dynamic graph learning for multi-scale traffic generation that incorporates spatial-temporal agents framework, allowing transportation participants using text to interact with real-time traffic. This AI-agent system facilitates the interaction between transportation participants and traffic data, addressing the concerns of transportation participants. First and foremost, we develop a text-to-demands agent to precisely extract traffic prediction tasks required by users, using a pre-prompted Q\&A agent to recognize needs from users' input. Coupled with our traffic prediction agent, prediction tasks are classified into three specific modules and output the results of future traffic conditions. Multi-scale traffic data are efficiently used in the traffic prediction agent, to generate multiple prediction results. Subsequently, the suggestion agent receives the real-time prediction results, to combination generate corresponding traffic suggestions for short-term or long-term planning. Finally, a scheme of visualization enhances the interactivity for the users, alleviating anxiety about the traffic. To validate the performance of our system, we conducted extensive qualitative and quantitative experiments on our self-collected clothing datasets. The results underscore the efficacy and
utility of each module within our system. 
In the future, we will explore more dimensional interactions and more accurate traffic prediction schemes to facilitate the formation of a universal transportation-human interaction paradigm.
\bibliographystyle{IEEEtran}
\bibliography{ARXIV_TRAFF}


%
%
%
%
%
%
%

\end{document}